\PassOptionsToPackage{table}{xcolor}
\documentclass[10pt,twocolumn,letterpaper]{article}
\usepackage[T1]{fontenc}

\usepackage[pagenumbers]{wacv} % To force page numbers, e.g. for an arXiv version

\definecolor{wacvblue}{rgb}{0.21,0.49,0.74}
\usepackage[pagebackref,breaklinks,colorlinks,allcolors=wacvblue]{hyperref}

\def\wacvPaperID{744} % *** Enter the WACV Paper ID here
\def\confName{WACV}
\def\confYear{2027}

\title{Spatial Transcriptomics Prediction from Histology via Topology-Informed Graph Attention and Gene-Semantics Conditioning}

\author{
Yuqi Cheng\textsuperscript{1}\qquad
Boshuai Zhang\textsuperscript{2}\qquad
Ziqi Zhang\textsuperscript{1}\qquad
Xiuwei Zhang\textsuperscript{1}\\[3pt]
\textsuperscript{1}Georgia Institute of Technology\qquad
\textsuperscript{2}University of California, San Diego\\[3pt]
{\ttfamily\small
cheng.yuqi@gatech.edu\quad
boz030@ucsd.edu\quad
ziqi.zhang@gatech.org\quad
xiuwei.zhang@gatech.edu}
}

\usepackage{xcolor}
\usepackage{colortbl}
\usepackage{multirow}

\definecolor{best}{RGB}{248,215,218}    % light red
\definecolor{second}{RGB}{207,226,255}  % light blue
\definecolor{groupgray}{gray}{0.92}     % group header background

\newcommand{\bestcell}[1]{\cellcolor{best}#1}
\newcommand{\secondcell}[1]{\cellcolor{second}#1}
\usepackage{algorithm}
\usepackage{algpseudocode}
\usepackage{dsfont}
\usepackage{amsmath,amssymb}

\newcommand{\grouprow}[2]{\rowcolor{groupgray}\multicolumn{#1}{l}{\textit{#2}}\\}

\begin{document}
\maketitle
\begin{abstract}
Spatial transcriptomics (ST) profiles gene expression with spatial context but remains costly, motivating models that infer expression from H\&E images. Existing approaches face three key limitations: (i) a single patch embedding is often insufficient—morphologically similar regions can exhibit distinct expression due to spatial context, while naive neighborhood aggregation can over-smooth boundaries; (ii) conventional fixed, gene-indexed output heads treat genes independently and cannot directly support strict zero-shot prediction for unseen genes; and (iii) common metrics under-measure spatial pattern fidelity and boundary resolution. To address these challenges, we propose TIGEM, a topology-informed, gene-conditioned framework for histology-to-ST inference. TIGEM learns topology-aware spot representations by regularizing graph attention with tissue-level topology and morphology cues to avoid cross-compartment smoothing, and uses gene semantic modulation to replace gene-indexed heads, enabling zero-shot prediction for unseen genes. We also introduce Graph Spectral Correlation (GSC) to evaluate spatial pattern fidelity and boundary resolution complement to PCC/MSE.  Across three large-scale ST datasets, TIGEM achieves overall state-of-the-art performance and significantly improved boundary prediction resolution, and further demonstrates the ability to predict expression patterns for genes unseen during training, extending prior zero-shot histology-to-ST efforts with topology-aware spatial propagation and PLM-based gene conditioning. The model is open-sourced at https://github.com/ZhangLabGT/TIGEM.
\end{abstract}
    
\vspace{-10pt}
\section{Introduction}
\label{sec:intro}

Spatial transcriptomics (ST) profiles gene expression while preserving tissue architecture, enabling links between molecular activity, morphology, and spatial organization~\cite{staahl2016visualization, larsson2021spatially}. Standard workflows partition hematoxylin and eosin (H\&E)-stained whole-slide images (WSI) into thousands of micrometer-scale spots and measure spot-level expression~\cite{moses2022museum, moffitt2022emerging}, providing insights into cellular heterogeneity and tissue microenvironments~\cite{rao2021exploring, oliveira2025high}. However, ST remains costly and operationally complex, limiting clinical deployment. In contrast, H\&E slides are inexpensive and routinely collected in clinical settings, motivating models that infer spatial gene expression directly from histology.

A common paradigm in histology-to-ST prediction is to extract a H\&E patch-level embedding for each spot and then regress spot-level expression~\cite{he2020integrating, zhu2025diffusion, yang2023exemplar, min2024multimodal}, optionally with graph-based neighborhood aggregation to incorporate spatial context~\cite{pang2021leveraging, chung2024accurate, shi2025multi, huang2025scalable, zeng2022spatial, huang2025stpath, zhao2026hist2cell}. Although this formulation appears straightforward, it relies on a strong implicit assumption: that local patch appearance (or together with all spatial neighbors) is sufficiently informative for downstream expression inference. In practice, however, this assumption is often invalid. Morphologically similar spots may still exhibit markedly different transcriptional programs when their surrounding tissue organization differs, for example across glandular lumens, compartment boundaries, stromal interfaces, or other microenvironmental transitions~\cite{zuo2024dissecting, elhanani2023spatial}. Some methods attempt to solve this problem by replacing deterministic regression with generative modeling and sampling-based prediction. Yet the core input assumption remains largely unchanged: the model is still asked to infer expression primarily from a single patch-level representation without explicitly modeling the surrounding tissue context.  Meanwhile, we should also noticed that naive neighborhood aggregation can mix distinct tissue contexts into overly similar representations, leading to over-smoothed predictions and reduced fidelity near spatial expression boundaries.

These observations suggest that the missing ingredient is not merely a stronger regressor, but a \emph{context-aware propagation rule} that adapts message passing to slide-specific tissue organization. In particular, spatial modeling should be guided by a learned notion of \emph{where information can plausibly travel} in the tissue plane, instead of assuming that spatial location proximity alone implies expression coupling.

A second limitation is that gene semantics remain underutilized in many histology-to-ST predictors. Conventional methods typically employ a fixed, gene-indexed output head of size $|G_{\text{train}}|$, restricting prediction to predefined training genes and preventing direct inference for unseen genes under a strict zero-shot split ($G_{\text{test}}\cap G_{\text{train}}=\emptyset$) without modifying or retraining the output layer. Recent studies\cite{yang2024spatial, cheng2026mixture} have begun to explore semantic conditioning for unseen-gene prediction, but they largely focus on gene-level generalization without explicitly modeling tissue topology or boundary-aware spatial propagation. This motivates gene-conditioned predictors that can transfer across genes while preserving spatial context.

Together, these limitations not only limit prediction resolution and boundary accuracy but also hinder clinical generalization. In addition, existing evaluation metrics are often overly simplistic, relying primarily on Pearson correlation. As a result, the spatial fidelity of a model at the boundary-level is often overlooked, and such metrics do not quantify the extent to which the predictions preserve biologically meaningful and spatially localized heterogeneity.

\begin{figure*}[h!]
\vspace{-15pt}
  \centering
   \includegraphics[width=0.8\linewidth]{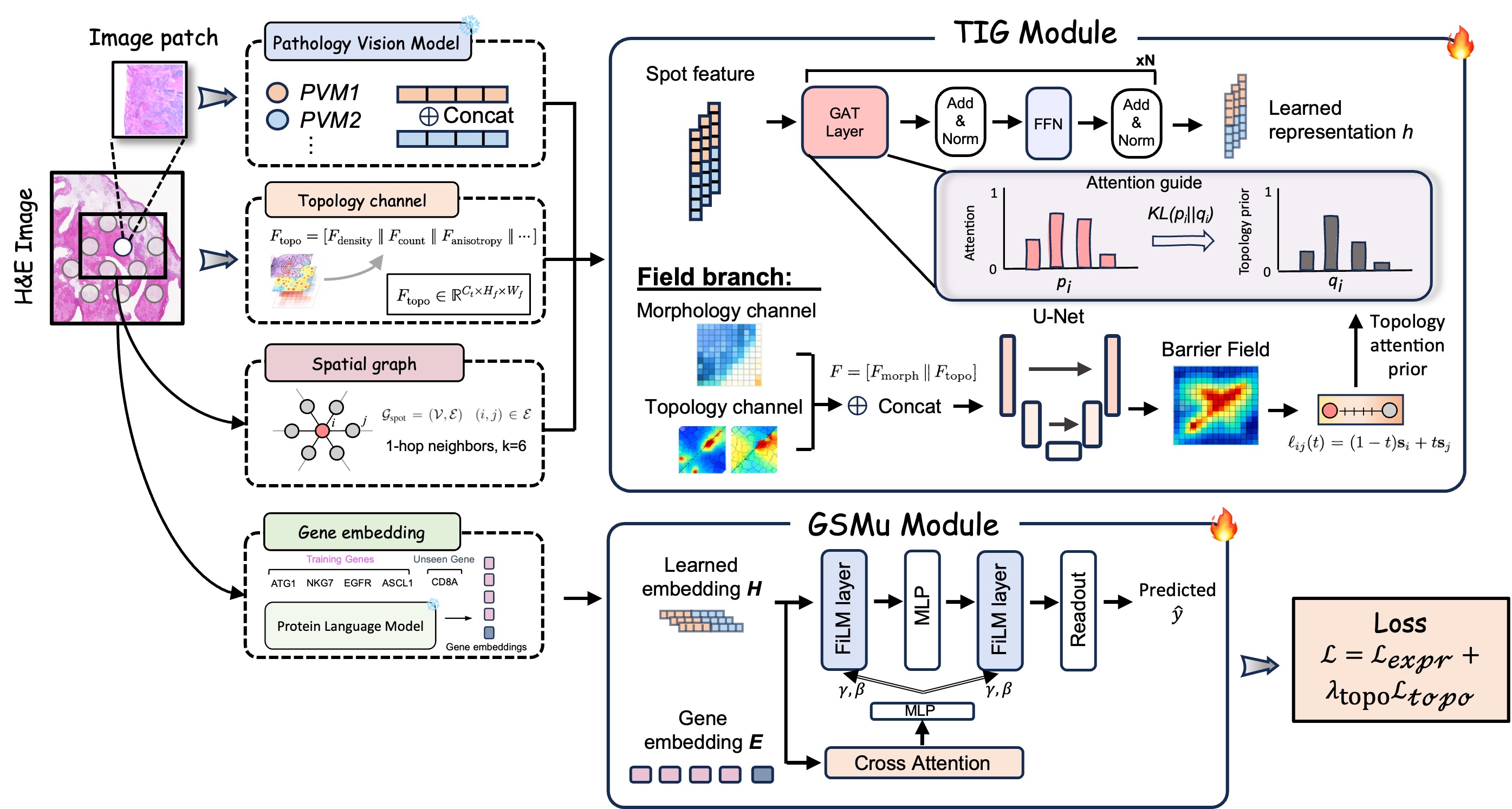}
   \caption{Overview of TIGEM. TIGEM includes two main modules: TIG module and GSMu module, enabling accurate spatial expression prediction and zero-shot gene inference.}
   \label{fig:one}
   \vspace{-15pt}
\end{figure*}

To address these limitations, we propose \textbf{TIGEM} (\textbf{T}opology-\textbf{I}nformed \textbf{G}raph attention with g\textbf{E}ne semantic \textbf{M}odulation), a framework for predicting both seen and unseen spatial gene expression from histology images (\cref{fig:one}). TIGEM first introduces a Topology-Informed Graph attention (TIG) module that learns a slide-specific continuous barrier field from morphology and topology cues to regularize graph attention, encouraging within-region propagation while suppressing message passing across potential tissue boundaries. TIGEM further replaces conventional gene-indexed output heads with a Gene Semantic Modulation unit (GSMu), where pretrained gene embeddings condition the readout and enable strict zero-shot inference by swapping in unseen-gene embeddings at test time. Together, these modules address both indiscriminate neighborhood aggregation and the fixed-output limitation of conventional predictors. We additionally introduce Graph Spectral Correlation (GSC), a graph-spectral metric that complements spot-wise measures such as PCC by evaluating spatial pattern fidelity. We provide its mathematical formulation and biological interpretation, and use it as a complementary criterion for benchmarking spatial expression patterns. We summarize our contributions as follows:

\begin{itemize}
    \item We present \textbf{TIGEM}, a topology-informed graph attention framework for histology-to-ST prediction, which significantly improves both overall expression accuracy and boundary fidelity, and achieves state-of-the-art results on three diverse ST benchmarks.

    \item By adopting pretrained pathology vision models (PVM) and protein language models (PLMs) \cite{chen2024uni, lu2024avisionlanguage, lin2022language} for the patch and gene embedding generation, TIGEM also significantly reduce the computational cost compared to existing prediction models.
    
    \item We extend zero-shot histology-to-ST prediction by introducing a PLM-conditioned GSMu decoder that supports strict unseen-gene inference, and combine it with topology-informed graph attention to improve spatial fidelity for clinically relevant biomarkers.
    
    \item We propose GSC, a graph-spectral metric that complements conventional magnitude-based measures (e.g., MSE and PCC) by quantifying spatial pattern preservation and boundary-resolving ability.
    
\end{itemize}

\vspace{-4pt}
\section{Related works}
\label{sec:rw}

\textbf{ST prediction from histology images.}
Predicting spatially resolved gene expression from H\&E-stained WSIs has attracted growing interest in computational pathology. Existing methods can fall into \emph{regression-based} and \emph{generative} paradigms. Regression approaches range from spot-wise predictors such as ST-Net~\cite{he2020integrating} to spatial-context models that aggregate neighborhood information and even multimodel information, such as HisToGene~\cite{pang2021leveraging}, Hist2ST~\cite{zeng2022spatial}, TRIPLEX~\cite{chung2024accurate}, and newly published M2TGLGO~\cite{shi2025multi}. Complementary lines improve robustness by aligning images and genes via contrastive-based designs (BLEEP~\cite{xie2023spatially}, EGN~\cite{yang2023exemplar}). Generative predictors are more recent, exemplified by STEM~\cite{zhu2025diffusion} using conditional diffusion. Despite these advances, most methods still instantiate a single mapping from morphology to expression and treat genes as independent targets, which can blur region-specific variability and limit generalization, especially to unseen genes.

\textbf{Zero-shot gene expression prediction.} Recent studies have begun to explore gene-level generalization beyond fixed gene-indexed prediction heads. SGN~\cite{yang2024spatial} introduces semantic gene information to enable zero-shot prediction of unseen genes from histology, while other methods such as M2TGLGO\cite{shi2025multi}, stImpute\cite {zeng2024imputing}, and HEIST~\cite {madhu2026heist} explore gene relationships, imputation, or gene-generalizable representation learning in spatial transcriptomics. These studies highlight the value of semantic or relational gene information, but differ substantially in task setting and input modality. TIGEM builds on this direction by combining PLM-based gene conditioning with topology-informed, boundary-aware graph propagation, jointly modeling unseen-gene transfer and spatial tissue context.

\textbf{Pathology vision models (PVMs).} Pathology vision models aim to learn general-purpose representations of histology image patches that can be transferred to downstream tasks such as gene expression prediction and prognosis analysis. Leveraging large-scale histology datasets, models such as UNI \cite{chen2024uni}, CONCH \cite{lu2024avisionlanguage}, Virchow\cite{vorontsov2024virchow}, and H-Optimus \cite{hoptimus0} are trained with self-supervised or multi-modal objectives. These foundation models capture rich morphological information that implicitly reflects cellular organization and molecular context, making them valuable for computational pathology and spatial omics applications.

\section{TIGEM}

\subsection{Preliminary}
\label{sec:3.1}

\textbf{Spatial transcriptomics and spots.} For a given slide, ST data provides a set of $N$ measurement spots $\mathcal{V} = \{1,\dots,N\}$ and $\mathbf{s}_i = (x_i,y_i)$ denotes the spot coordinate in the WSI. Around each spot $i$, a fixed-size image patch $I_i \in \mathbb{R}^{H \times W \times 3}$ ($224\times224$ pixels) is cropped from the H\&E slide. 
Benefiting from pretrained pathology vision models (PVMs), we can directly use their patch embeddings instead of extracting features from scratch, substantially improving computational efficiency. We assume a pretrained PVM $f_{PVM}(\cdot)$  maps $I_i$ to a feature vector $\mathbf{v}_i \in \mathbb{R}^{d_v}$, following
\begin{equation}
    \mathbf{v}_i = f_{PVM}(I_i).
\end{equation}
In our experiments, we benchmark multiple PVM backbones and their combinations, and use the best-performing combination UNI-2\cite{chen2024uni}+Virchow-2\cite{vorontsov2024virchow} as $\mathbf{v}_i$. We further use a trainable graph encoder (GE) $f_{GE}$ to encode neighborhood information into the primary morphology feature $\mathbf v_i$, obtaining the spot histology representation $\mathbf h_i$.
\begin{equation}
    \mathbf{h}_i = f_{GE}(\mathbf v_i).
\end{equation}

For a set of $G$ genes $\mathcal{G} = \{g_1,\dots,g_G\}$, we denote the observed expression at spot $i$ as $ \mathbf{y}_i = \big(y_{i,g}\big)_{g \in \mathcal{G}} \in \mathbb{R}^G$, where $y_{i,g}$ is the log-normalized gene expression value.

\textbf{Gene embedding.} For each gene, we use its corresponding protein embedding $\mathbf{e}_g \in \mathbb{R}^{d_g},$ obtained from the pretrained protein language model $f_{PLM}(\cdot)$ following
\begin{equation}
    \mathbf{e}_g = f_{PLM}(g_{\text{protein}}).
\end{equation}
In practice, we use ESM2\cite{lin2023evolutionary} to generate $\mathbf e_g$ from the amino-acid sequence of gene $g$.

\subsection{Topology-Informed GAT (TIG) Module}
\label{sec:tig_block}

The goal of the TIG module (\cref{fig:one}) is to learn boundary-aware spot representations on a spatial graph.
Because tissue architecture is non-homogeneous, we adopt graph attention (GAT) to enable task-driven neighbor weighting.
However, standard GAT still normalizes attention over geometric neighbors using only patch-derived features, which can lead to information leakage across compartment boundaries and thus over-smoothed predictions near expression transitions.
To address this, we introduce a global \emph{barrier-field branch} that estimates a slide-specific barrier map and uses it to regularize attention, discouraging cross-boundary propagation.

\textbf{Field branch for continuous barrier estimation.}
We learn a continuous barrier field over the tissue plane from fused morphology--topology cues.
We first construct a dense spatial feature map
\begin{equation}
F = [F_{\mathrm{morph}} \,\|\, F_{\mathrm{topo}}] \in \mathbb{R}^{C \times H_f \times W_f},
\end{equation}
aligned with slide coordinates, and predict a nonnegative barrier map via a lightweight U-Net,
\begin{equation}
B = f_{\mathrm{field}}(F) \in \mathbb{R}^{H_f \times W_f},
\end{equation}
where larger values indicate regions through which propagation is less plausible.

\textit{Morphology channels.}
Given spot embeddings $\mathbf{x}_i \in \mathbb{R}^{d_x}$ and spot coordinates $\mathbf{s}_i$, we project each embedding into a lower-dimensional morphology space,
\begin{equation}
\tilde{\mathbf{x}}_i = W_x \mathbf{x}_i \in \mathbb{R}^{d_m},
\end{equation}
and rasterize these projected features onto a 2D grid using Gaussian splatting:
\begin{equation}
F_{\mathrm{morph}}(\mathbf{u})
=
\sum_{i=1}^{N}
K_{\sigma}(\mathbf{u}-\mathbf{s}_i)\,\tilde{\mathbf{x}}_i,
\end{equation}
where $\mathbf{u}$ is a grid location and $K_{\sigma}$ is a Gaussian kernel with bandwidth $\sigma$.

\textit{Topology channels.}
To further incorporate spatial structural information beyond patch appearance, we construct an additional topology cue map $F_{\mathrm{topo}}$. We use CellViT\cite{horst2024cellvit} (which is the default in HEST-1k dataset \cite{jaume2024hest}) to perform cell segmentation for the slide. In practice, this map includes processed cell- or tissue-level descriptors such as cell density, local nuclei count, and anisotropy statistics extracted from the histology image. These channels provide complementary evidence about local connectivity, discontinuity, and compartment transitions that are not fully captured by morphology embeddings alone. The detailed constructions of the topology descriptors are provided in the \textit{Supplementary Material}.

\begin{equation}
    F_{\mathrm{topo}}
=
\big[
F_{\text{density}}
\;\|\;
F_{\text{count}}
\;\|\;
F_{\text{anisotropy}}
\;
\big]
\end{equation}

\textit{Barrier cost.} For an edge $(i,j)\in\mathcal E$, let $\ell_{ij}(t)=(1-t)\mathbf s_i+t\mathbf s_j$, $t\in[0,1]$, denote the line segment connecting the two spots. We sample $T$ points uniformly along this segment and define the traversal cost
\begin{equation}
    c_{ij} =
    \frac{\|\mathbf s_i-\mathbf s_j\|_2}{T}
    \sum_{t=1}^{T}
    B\!\left(\ell_{ij}(\tau_t)\right),
\end{equation}
where $\{\tau_t\}_{t=1}^{T}$ are uniformly spaced samples in $[0,1]$. 

\textit{Pairwise supervision for training the barrier field.}
The field branch is pre-trained with an independent weak pairwise objective defined on edges.
We construct \emph{positive} edges $\mathcal P$ that are likely within the same compartment and \emph{negative} edges $\mathcal N$ that likely cross a boundary (based on morphology/topology cues; see \textit{Supplementary}).
The barrier field is trained to assign smaller traversal costs to positives than to negatives using a margin ranking loss:
\begin{equation}
\mathcal L_{\mathrm{field}}
=
\frac{1}{|\mathcal P|}
\sum_{(i,j)\in\mathcal P}
\frac{1}{|\mathcal N(i)|}
\sum_{(i,k)\in\mathcal N(i)}
\max\!\big(0,\; m + c_{ij} - c_{ik}\big),
\label{eq:loss_field}
\end{equation}
where $m>0$ is a margin and $\mathcal N(i)=\{k\mid(i,k)\in\mathcal N\}$.
This prevents degenerate solutions (e.g., a flat $B$) and provides a direct learning signal for updating $f_{\mathrm{field}}$.

Finally, the barrier-induced distribution over the neighborhood of node $i$ is then derived from the pre-trained fixed $c_{ij}$
\begin{equation}
    q_{ij}
    =
    \mathrm{softmax}_{j\in\mathcal N(i)}
    \left(
    -\frac{c_{ij}}{\tau_b}
    \right),
    \label{eq:teacher_dist}
\end{equation}
where $\tau_b$ is a temperature parameter.

\textbf{Topology-informed graph attention.}
Given node features $\mathbf h_i^{(l-1)}$ from layer $(l-1)$, a multi-head graph attention layer computes attention logits
\begin{equation}
e_{ij}^{(m,l)}
=
\mathrm{Attn}^{(m,l)}\!\left(\mathbf h_i^{(l-1)},\mathbf h_j^{(l-1)}\right),
\end{equation}
for head $m$ and neighbor $j\in\mathcal N(i)$. TIGEM targets Visium-style layouts, where each spot has a canonical first-ring neighborhood of 6 hexagonal neighbors. Thus $\mathcal N(i)$ is the surrounding 6 nearest neighbors (\cref{fig:one}). The corresponding attention distribution is
\begin{equation}
p_{ij}^{(m,l)}
=
\mathrm{softmax}_{j\in\mathcal N(i)}\!\left(e_{ij}^{(m,l)}\right).
\end{equation}
We use the average attention across heads as
\begin{equation}
    \bar p_{ij}^{(l)} = \frac{1}{M}\sum_{m=1}^{M} p_{ij}^{(m,l)}.
\end{equation}
The topology-informed aggregation at layer $l$ is given by
\begin{equation}
\tilde{\mathbf h}_i^{(l)}
=
\mathop{\Vert}_{m=1}^{M}
\sum_{j\in\mathcal N(i)}
p_{ij}^{(m,l)}\, W_V^{(m,l)} \mathbf h_j^{(l-1)},
\end{equation}
where $\Vert$ denotes concatenation over heads. Each TIG block further includes a residual connection, layer normalization, and a position-wise feed-forward network:
\begin{align}
\hat{\mathbf h}_i^{(l)} &=
\mathrm{LN}\!\left(
\mathbf h_i^{(l-1)} + \tilde{\mathbf h}_i^{(l)}
\right),\\
\mathbf h_i^{(l)} &=
\mathrm{LN}\!\left(
\hat{\mathbf h}_i^{(l)} + \mathrm{FFN}^{(l)}(\hat{\mathbf h}_i^{(l)})
\right).
\end{align}
Stacking $L$ TIG blocks yields the final node embedding matrix
\begin{equation}
H=[\mathbf h_1,\dots,\mathbf h_N]^\top \in \mathbb{R}^{N\times d}.
\end{equation}

\textbf{Topology supervision.}
The learned barrier field does not directly predict expression. Instead, it provides a spatially structured distribution that regularizes graph attention. We use a KL divergence between the field-induced distribution in Eq.~\eqref{eq:teacher_dist} and the GAT attention distribution:
\begin{equation}
\mathcal L_{\mathrm{topo}}
=
\frac{1}{L}\sum_{l=1}^{L}
\frac{1}{N}\sum_{i=1}^{N}
\mathrm{KL}\!\left(
\mathrm{stopgrad}(q_i)\;\|\;\bar p_i^{(l)}
\right).
\label{eq:loss_topo}
\end{equation}
Importantly, gradients are stopped through $q$ (and thus $B$) in $\mathcal L_{\mathrm{topo}}$, so this term updates only the GAT encoder and prevents the field branch from trivially adapting to match the GAT attention.

\subsection{Gene Semantic Modulation unit (GSMu)}
\vspace{-3pt}
\label{sec:cafilm-head}

Given the learned node embeddings $H \in \mathbb{R}^{N \times d}$ from the TIG encoder and gene embeddings $E \in \mathbb{R}^{G \times d_g}$ from ESM2, our goal is to perform gene-conditioned expression readout.
Instead of learning $G$ independent prediction heads, we use the gene embeddings to modulate a shared morphological representation, so that each gene is decoded as\textit{ this tissue pattern, as seen from the perspective of this gene}.
Herein, we design a GSMu module that combines cross-attention and feature-wise linear modulation (FiLM)-style conditioning~\cite{perez2018film} (\cref{fig:one}).

\textbf{Gene tokens and cross-attention.} We first normalize and linearly project each gene embedding $\mathbf{e}_g$ from ESM2 to the encoder dimension, obtaining gene tokens $\phi_g \in \mathbb{R}^d$, $g=1,\dots,G$.
A single cross-attention layer then lets each gene token ``look at'' the tissue morphology: gene tokens act as queries, and the node embeddings $H$ provide keys and values,
\begin{equation}
\mathbf{z}_g^{\text{gene}}
=
\mathrm{CrossAttn}\big(\phi_g, H\big)
\in \mathbb{R}^d,
\label{eq:gene-ca-brief}
\end{equation}
so that $\mathbf{z}_g^{\text{gene}}$ becomes a morphology-aware representation of gene $g$.
This step explicitly aligns sequence-level semantics with slide-specific histology before decoding expression.

\textbf{FiLM modulation for gene-specific readout.} From each refined gene token $\mathbf{z}_g^{\text{gene}}$ we generate FiLM parameters via a small MLP, producing a channel-wise scaling and shift $(\boldsymbol{\gamma}_g, \boldsymbol{\beta}_g) \in \mathbb{R}^d \times \mathbb{R}^d$.
Given a node embedding $\mathbf{h}_i \in \mathbb{R}^d$, we obtain a gene-specific modulated feature
\begin{equation}
\tilde{\mathbf{h}}_{i,g}
=
\boldsymbol{\gamma}_g \odot \mathbf{h}_i
+
\boldsymbol{\beta}_g,
\label{eq:film-brief}
\end{equation}
and apply a shared readout MLP to predict expression:
\begin{equation}
\hat y_{i,g}
=
f_{\text{head}}\!\left(\tilde{\mathbf h}_{i,g}\right).
\label{eq:gsmu_pred}
\end{equation}
This removes gene-indexed parameters from the output and instead decodes expression through semantic conditioning.

\textbf{Zero-shot prediction for unseen genes.}
Because GSMu is fully conditioned on gene embeddings (rather than gene indices), TIGEM supports strict zero-shot inference.
At test time, for an unseen gene $g^\ast$ we compute its embedding $\mathbf e_{g^\ast}$, form the token $\phi_{g^\ast}$, and reuse the same cross-attention and FiLM-conditioned readout to predict $\hat y_{i,g^\ast}$. The topology-informed encoder remains unchanged.

\subsection{Total Training Objective}
\vspace{-3pt}
\label{sec:total-loss}

We train the model to predict log-transformed expression for spot $i$ and gene $g$.
The main supervision is a mean squared error (MSE) over all observed spots and genes. In addition, we use the topology-guided KL term from Section~\ref{sec:tig_block} to align attention distributions with the topology prior. The overall loss is
\begin{equation}
\mathcal{L}
=
\mathcal{L}_{\text{MSE}}
+
\lambda_{\text{topo}}
\mathcal{L}_{\text{topo}},
\label{eq:total-loss}
\end{equation}
where $\lambda_{\text{topo}}$ controls the strength of topology guidance.
\section{Experiments}
\vspace{-4pt}
\subsection{Datasets}
\vspace{-4pt}
We selected three publicly available spatial transcriptomics datasets, named HEART, COAD, and MEND datasets, to do the benchmarking. Specifically, HEART \cite{kanemaru2023spatially} consists of 9 human heart tissue sections from independent samples, COAD \cite{heiser2023molecular} contains 13 human colorectal cancer samples, and MEND \cite{erickson2022spatially} includes 14 human prostate cancer sections. These datasets are also available in HEST-1k\cite{jaume2024hest}.

\vspace{-2pt}
\subsection{Experiment Setting}
\vspace{-4pt}

All H\&E slides are cropped into $224\times224$ patches centered at each spot. For all benchmark comparisons, TIGEM analyses, and ablation studies, we adopt leave-one-out cross-validation (LOOCV) over all slides. In each fold, one slide is held for testing and the remaining slides are used for training. We report the mean and standard deviation across folds.

Across baselines, we select the top 500 genes from the intersection of highly expressed and highly variable genes (HEHVG) as targets, and apply log1p to all counts. Hyperparameters are reported in the \textit{Supplementary}; we initialize from official settings and tune key hyperparameters on the validation split for fair comparison.

\begin{table*}[h]
  \caption{Benchmark results of TIGEM, existing ST prediction models, and PVM baselines (Group 1). Best and second-best results in each column are highlighted with red and blue backgrounds, respectively. Higher is better for PCC(50), PCC(200), GSC@90\%, and SSIM (50), while lower is better for MSE. All values are reported as mean with standard deviation shown in the subscript.}
  \label{table:bench-merged-tigem}
  \centering
  \large
  \setlength{\tabcolsep}{6pt}
  \renewcommand{\arraystretch}{1}
  \vspace{-7pt}
  \resizebox{\textwidth}{!}{%
  \begin{tabular}{l *{15}{c}}
    \toprule
    Model
      & \multicolumn{5}{c}{HEART (LOOCV)}
      & \multicolumn{5}{c}{COAD (LOOCV)}
      & \multicolumn{5}{c}{MEND (LOOCV)} \\
    \cmidrule(lr){2-6}\cmidrule(lr){7-11}\cmidrule(lr){12-16}
      & MSE $\downarrow$ & PCC(50) $\uparrow$ & PCC(200) $\uparrow$ & GSC@90\% $\uparrow$ & SSIM (50) $\uparrow$
      & MSE $\downarrow$ & PCC(50) $\uparrow$ & PCC(200) $\uparrow$ & GSC@90\% $\uparrow$ & SSIM (50) $\uparrow$
      & MSE $\downarrow$ & PCC(50) $\uparrow$ & PCC(200) $\uparrow$ & GSC@90\% $\uparrow$ & SSIM (50) $\uparrow$ \\
    \midrule

    % ---------------- group: ST prediction methods ----------------
    \grouprow{16}{ST prediction methods}

    ST-Net~\cite{he2020integrating}
      & $0.365_{.074}$ & $0.208_{.083}$ & $0.109_{.027}$ & $0.054_{.015}$ & $0.063_{.013}$
      & $1.078_{.093}$ & $0.574_{.055}$ & $0.448_{.044}$ & $0.402_{.053}$ & $0.359_{.045}$
      & $0.333_{.029}$ & $0.577_{.056}$ & $0.428_{.030}$ & $0.338_{.026}$ & $0.364_{.038}$\\

    HistoGene~\cite{pang2021leveraging}
      & $0.468_{.055}$ & $0.159_{.029}$ & $0.082_{.032}$ & $0.044_{.012}$ & $0.049_{.009}$
      & $1.019_{.094}$ & $0.561_{.063}$ & $0.470_{.058}$ & $0.452_{.046}$ & $0.413_{.033}$
      & $0.354_{.030}$ & $0.387_{.035}$ & $0.238_{.038}$ & $0.176_{.032}$ & $0.202_{.019}$ \\

    THItoGene~\cite{jia2023thitogene}
      & $0.528_{.088}$ & $0.125_{.048}$ & $0.098_{.022}$ & $0.041_{.014}$ & $0.040_{.011}$
      & $1.893_{.101}$ & $0.528_{.050}$ & $0.400_{.045}$ & $0.288_{.050}$ & $0.231_{.042}$
      & $0.578_{.042}$ & $0.278_{.039}$ & $0.179_{.029}$ & $0.101_{.039}$ & $0.135_{.022}$ \\

    BLEEP~\cite{xie2023spatially}
      & $0.327_{.070}$ & $0.276_{.049}$ & $0.198_{.034}$ & $0.101_{.016}$ & $0.103_{.018}$
      & $0.992_{.098}$ & $0.577_{.062}$ & $0.483_{.040}$ & $0.427_{.036}$ & $0.360_{.041}$
      & $0.361_{.031}$ & $0.594_{.044}$ & $0.439_{.037}$ & $0.329_{.022}$ & $0.350_{.035}$ \\

    TRIPLEX~\cite{chung2024accurate}
      & $0.545_{.057}$ & $0.295_{.051}$ & $0.212_{.048}$ & $0.114_{.014}$ & $0.117_{.019}$
      & \secondcell{$0.726_{.083}$} & $0.697_{.067}$ & $0.598_{.049}$ & $0.553_{.049}$ & $0.515_{.035}$
      & $0.318_{.024}$ & $0.669_{.036}$ & $0.529_{.033}$ & $0.399_{.030}$ & $0.428_{.036}$ \\

    Stem~\cite{zhu2025diffusion}
      & $0.503_{.050}$ & $0.224_{.045}$ & $0.114_{.022}$ & $0.089_{.021}$ & $0.086_{.013}$
      & $1.387_{.116}$ & $0.567_{.064}$ & $0.467_{.056}$ & $0.507_{.036}$ & $0.449_{.047}$
      & $0.537_{.033}$ & $0.574_{.041}$ & $0.388_{.029}$ & $0.336_{.036}$ & $0.382_{.029}$ \\

    istar~\cite{zhang2024inferring}
      & $0.321_{.058}$ & $0.252_{.040}$ & $0.154_{.032}$ & $0.123_{.018}$ & $0.120_{.020}$
      & $0.897_{.076}$ & $0.664_{.055}$ & $0.604_{.041}$ & $0.541_{.051}$ & $0.487_{.035}$
      & $0.435_{.032}$ & $0.638_{.050}$ & $0.484_{.045}$ & $0.360_{.026}$ & $0.392_{.030}$ \\

    M2TGLGO~\cite{ganguly2025merge}
      & $0.328_{.044}$ & $0.292_{.048}$ & $0.155_{.037}$ & $0.120_{.025}$ & $0.114_{.016}$
      & $0.874_{.077}$ & $0.672_{.068}$ & $0.583_{.055}$ & $0.530_{.043}$ & $0.493_{.054}$
      & $0.376_{.036}$ & $0.609_{.052}$ & $0.497_{.043}$ & $0.371_{.045}$ & $0.409_{.034}$ \\

    mclSTExp~\cite{min2024multimodal}
      & $\secondcell{0.297_{.037}}$ & $0.236_{.032}$ & $0.144_{.026}$ & $0.118_{.013}$ & $0.113_{.017}$
      & $0.913_{.079}$ & $0.611_{.048}$ & $0.523_{.056}$ & $0.497_{.036}$ & $0.448_{.034}$
      & \secondcell{$0.316_{.028}$} & $0.602_{.033}$ & $0.472_{.030}$ & $0.348_{.036}$ & $0.376_{.050}$ \\

    STFlow~\cite{huang2025scalable}
      & $0.324_{.041}$ & $0.281_{.039}$ & $0.149_{.030}$ & $0.113_{.017}$ & $0.114_{.016}$
      & $0.861_{.082}$ & $0.669_{.071}$ & $0.589_{.043}$ & \secondcell{$0.558_{.041}$} & \secondcell{$0.522_{.042}$}
      & $0.322_{.036}$ & \secondcell{$0.687_{.048}$} & $\secondcell{0.534_{.036}}$ & $\secondcell{0.411_{.035}}$ & \secondcell{$0.447_{.039}$} \\

    MERGE~\cite{ganguly2025merge}
      & $0.355_{.049}$ & $\secondcell{0.358_{.051}}$ & $\secondcell{0.221_{.043}}$ & $\secondcell{0.143_{.015}}$ & \secondcell{$0.149_{.023}$}
      & $0.813_{.086}$ & $\secondcell{0.705_{.059}}$ & $\secondcell{0.619_{.043}}$ & $0.538_{.040}$ & $0.477_{.030}$
      & $0.424_{.032}$ & $0.674_{.050}$ & $0.507_{.039}$ & $0.359_{.047}$ & $0.387_{.029}$ \\

    \midrule

    % ---------------- group: PVM encoders ----------------
    \grouprow{16}{PVM encoders}

    Virchow2~\cite{vorontsov2024virchow}
      & $0.366_{.033}$ & $0.258_{.035}$ & $0.181_{.039}$ & $0.065_{.013}$ & $0.069_{.008}$
      & $0.907_{.068}$ & $0.562_{.047}$ & $0.482_{.039}$ & $0.468_{.045}$ & $0.417_{.028}$
      & $0.327_{.027}$ & $0.613_{.039}$ & $0.495_{.033}$ & $0.334_{.032}$ & $0.374_{.034}$ \\

    CONCH~\cite{lu2024avisionlanguage}
      & $0.380_{.024}$ & $0.189_{.041}$ & $0.106_{.018}$ & $0.042_{.010}$ & $0.040_{.010}$
      & $1.125_{.118}$ & $0.505_{.042}$ & $0.395_{.046}$ & $0.337_{.042}$ & $0.266_{.033}$
      & $0.346_{.035}$ & $0.541_{.027}$ & $0.427_{.041}$ & $0.297_{.028}$ & $0.339_{.035}$ \\

    UNI-2~\cite{chen2024uni}
      & $0.348_{.044}$ & $0.243_{.032}$ & $0.205_{.038}$ & $0.093_{.016}$ & $0.097_{.013}$
      & $0.922_{.077}$ & $0.581_{.063}$ & $0.476_{.044}$ & $0.438_{.031}$ & $0.364_{.038}$
      & $0.324_{.025}$ & $0.608_{.037}$ & $0.478_{.026}$ & $0.328_{.031}$ & $0.369_{.041}$ \\

    GigaPath~\cite{xu2024whole}
      & $0.351_{.042}$ & $0.218_{.029}$ & $0.123_{.036}$ & $0.072_{.012}$ & $0.075_{.018}$
      & $0.997_{.084}$ & $0.537_{.044}$ & $0.428_{.030}$ & $0.417_{.038}$ & $0.355_{.028}$
      & $0.330_{.039}$ & $0.550_{.046}$ & $0.447_{.032}$ & $0.288_{.037}$ & $0.322_{.023}$ \\

    UNI-2 + Virchow2
      & $0.328_{.037}$ & $0.277_{.033}$ & $0.212_{.031}$ & $0.098_{.016}$ & $0.103_{.019}$
      & $0.904_{.083}$ & $0.585_{.063}$ & $0.496_{.038}$ & $0.480_{.038}$ & $0.437_{.029}$
      & $0.320_{.024}$ & $0.624_{.044}$ & $0.499_{.036}$ & $0.346_{.047}$ & $0.388_{.037}$\\

    CONCH + UNI-2
      & $0.349_{.054}$ & $0.228_{.035}$ & $0.203_{.040}$ & $0.091_{.013}$ & $0.088_{.017}$
      & $1.007_{.093}$ & $0.548_{.052}$ & $0.468_{.039}$ & $0.387_{.040}$ & $0.340_{.024}$
      & $0.337_{.032}$ & $0.600_{.039}$ & $0.467_{.028}$ & $0.325_{.039}$ & $0.362_{.044}$ \\

    CONCH + UNI-2 + Virchow2
      & $0.334_{.033}$ & $0.264_{.037}$ & $0.215_{.032}$ & $0.102_{.015}$ & $0.106_{.018}$
      & $0.972_{.080}$ & $0.599_{.043}$ & $0.503_{.048}$ & $0.491_{.042}$ & $0.450_{.031}$
      & $0.322_{.030}$ & $0.634_{.057}$ & $0.495_{.027}$ & $0.345_{.028}$ & $0.379_{.029}$\\

    \midrule

    % ---------------- group: Ours ----------------
    \grouprow{16}{Ours}

    \textbf{TIGEM}
      & \bestcell{$0.245_{.049}$} & \bestcell{$0.388_{.064}$} & \bestcell{$0.241_{.049}$} & \bestcell{$0.160_{.015}$} & \bestcell{$0.175_{.020}$}
      & \bestcell{$0.687_{.074}$} & \bestcell{$0.718_{.061}$} & \bestcell{$0.631_{.043}$} & \bestcell{$0.584_{.034}$} & \bestcell{$0.539_{.046}$}
      & \bestcell{$0.308_{.043}$} & \bestcell{$0.701_{.038}$} & \bestcell{$0.549_{.045}$} & \bestcell{$0.422_{.038}$} & \bestcell{$0.466_{.054}$} \\

    \bottomrule
  \end{tabular}}
  \vspace{-10pt}
\end{table*}

\vspace{-2pt}
\subsection{Evaluation Metric}
\label{sec:evm}
\vspace{-4pt}

We report mean squared error (MSE) and Pearson correlation coefficient (PCC), following prior work \cite{zhu2025diffusion, chung2024accurate}. PCC is evaluated at two scales, PCC(50) and PCC(200), calculating the average of top 50 and top 200 PCC values. 

Meanwhile, we also notice that MSE and PCC primarily measure vector-level agreement and may overlook spatial fidelity. In practice, predicted expression maps are often smoother than the ground truth~\cite{zhu2025diffusion, xie2023spatially}, making PCC less sensitive to gene spatial pattern mismatches, while MSE is dominated by magnitude differences and does not explicitly assess spatial structure. To explicitly evaluate spatial pattern fidelity, we design \textbf{Graph Spectral Correlation (GSC)} in collaboration with biological experts. GSC is a graph signal processing metric that compares predictions and ground truth in the graph Fourier domain. Given the spot adjacency graph with the normalized Laplacian
$L = I - D^{-1/2} A D^{-1/2}$ and eigenbasis $U$,
we compute the cosine similarity between the truncated spectral coefficients that retain a proportion $p$ of cumulative spectral energy:
\begin{equation}
\mathrm{GSC}(p)=
\frac{\langle \hat{x}_{\mathrm{true}}^{(p)},\,\hat{x}_{\mathrm{pred}}^{(p)}\rangle}{
\|\hat{x}_{\mathrm{true}}^{(p)}\|_2\,\|\hat{x}_{\mathrm{pred}}^{(p)}\|_2+\varepsilon }.
\end{equation}
We report GSC@90\% in the main results as a resolution-oriented metric. Take \cref{fig:four} as a justification, GSC@90\% faithfully measures spatial pattern than PCC. Although Stem model achieves noticeably higher PCC than ST-Net and BLEEP, its predictions are overly smoothed, resulting in blurred boundaries and failure to recover the spatial pattern of KLK3. Thus it does not exhibit a corresponding advantage under GSC@90\%, highlighting the metric’s ability to penalize spatial mismatches that PCC overlooks.
A complete mathematical proof and biological interpretation of GSC are provided in the \textit{Supplementary Material}. 

To provide an independent assessment of spatial fidelity, we additionally report Structural Similarity Index (SSIM) with top 50 values SSIM (50), an established image-structure metric that is independent of both GSC and the TIGEM training objective.

\subsection{ST Expression Prediction}
\label{sec:stpred}

\textbf{Comparing with H\&E-based ST prediction models.}
We first benchmarked our method with 11 published ST prediction models. As shown in \cref{table:bench-merged-tigem}, TIGEM consistently outperforms all baseline models across all evaluation metrics and achieves state-of-the-art performance on three datasets spanning distinct tissue types and pathological conditions. Benefiting from the incorporation of cell-level geometric topology, TIGEM more effectively preserves spatial expression structure, leading to substantially higher GSC@90\% scores compared with competing methods. Importantly, TIGEM also achieves the highest SSIM on all three datasets, independently corroborating the spatial improvements observed with GSC@90\% using an established spatial-structure metric. Since all histology sections originate from different patients or sequencing batches, these results also suggest that TIGEM is robust to technical noise and generalizes well to unseen query slices.

\textbf{Evaluated TIGEM against recent pathology vision models (PVMs)}. To assess whether TIGEM performance comes merely from strong histology embeddings, we included several open-source PVMs (Virchow2, UNI-2, CONCH, and GigaPath) as well as their combinations. We followed the HEST-1K protocol by training a random forest regressor on their patch embeddings. Across all datasets, TIGEM still surpasses these models and their combinations on all metrics, with a particularly large margin on the spatial-pattern metric GSC@90\% 
(\cref{table:bench-merged-tigem}). This demonstrates that TIGEM predicts expression in a higher resolution than using PVM embeddings alone, and that its framework substantially enhances the spatial organization awareness of existing pathology encoders.

\vspace{-4pt}
\subsection{Biomarker Recovery}
\label{sec:visual}

\begin{figure*}[h]
\vspace{-10pt}
  \centering
  \includegraphics[width=1\linewidth]{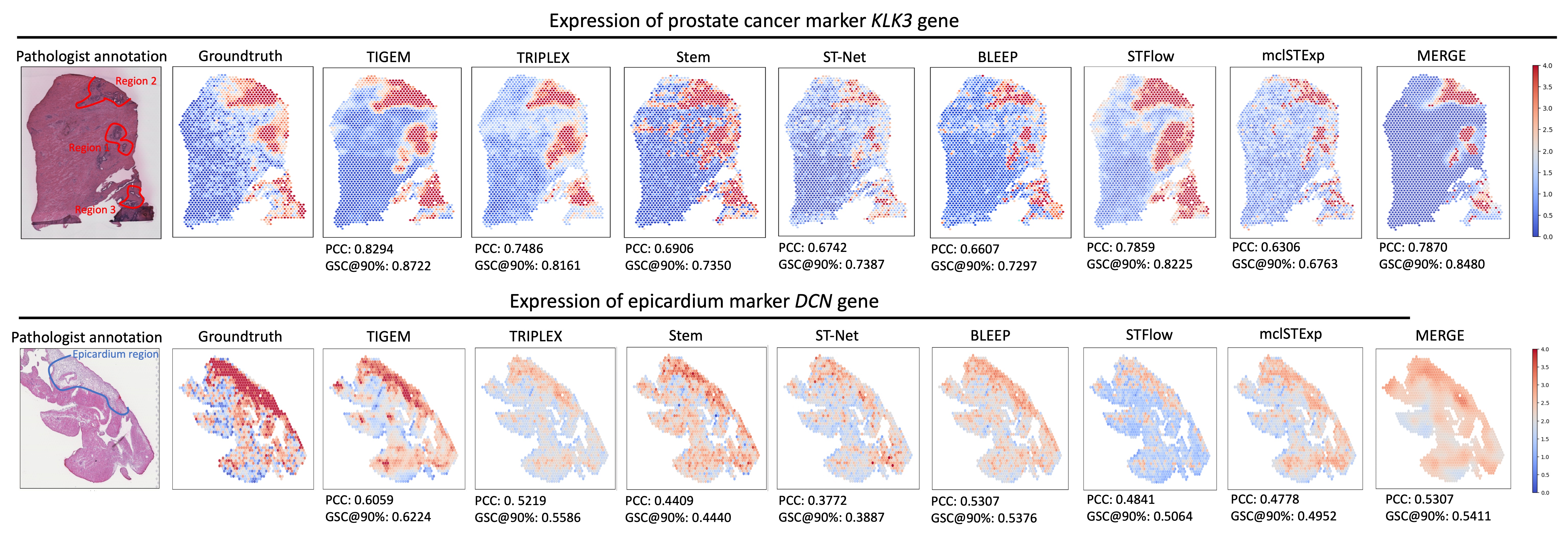}
  \caption{Visualization of biomarker expression prediction across methods. Top shows the prediction of prostate cancer marker KLK3 on the MEND dataset. Region 1-3 are tumor regions. Bottom shows the prediction of epicardium marker DCN on the HEART dataset.}
  \label{fig:four}
  \vspace{-10pt}
\end{figure*}

One key application of spatial gene expression prediction is to uncover the relationship between disease progression and tissue morphology, which requires accurate biomarker estimation. We therefore evaluated TIGEM and the top eight baseline methods on four biomarker genes across both healthy H\&E sections (HEART dataset) and cancer tissue (MEND dataset, prostate cancer). In MEND, we selected two established prostate cancer markers, KLK3\cite{penney2011association, srinivasan2024psa} and AZGP1\cite{wen2024azgp1}; in HEART, we evaluated the cardiomyocyte marker ANKRD1\cite{arimura2009cardiac} and the epicardium marker DNC\cite{yan2009decorin}. As shown in \cref{fig:four}, TIGEM’s predictions exhibit strong alignment with the ground truth and closely overlap with pathology-annotated regions, producing sharply defined expression boundaries. These results demonstrate TIGEM’s strong ability to capture spatially coherent biomarker patterns under topology-informed guidance. Additional visualizations are shown in \textit{Supplementary Material}.

\begin{table*}[h]
\vspace{5pt}
  \caption{Zero-shot prediction performance of TIGEM under different unseen gene settings. Best and second-best results in each column are highlighted with red and blue backgrounds, respectively.}
  \label{tab:zeroshot_merged}
  \centering
  \small
  \renewcommand{\arraystretch}{0.95}
  \setlength{\tabcolsep}{5.5pt}

  \resizebox{\textwidth}{!}{%
  \begin{tabular}{@{}llccc|ccc|ccc@{}}
    \toprule
    \multirow{2}{*}{Setting} & \multirow{2}{*}{Ratio}
      & \multicolumn{3}{c|}{HEART (MISC142)}
      & \multicolumn{3}{c|}{COAD (MISC36)}
      & \multicolumn{3}{c}{MEND (MEND146)} \\
    \cmidrule(lr){3-5}\cmidrule(lr){6-8}\cmidrule(lr){9-11}
      & & MSE $\downarrow$ & PCC(all) $\uparrow$ & GSC@90\% $\uparrow$
        & MSE $\downarrow$ & PCC(all) $\uparrow$ & GSC@90\% $\uparrow$
        & MSE $\downarrow$ & PCC(all) $\uparrow$ & GSC@90\% $\uparrow$ \\
    \midrule

    \multirow{3}{*}{HEHVG}
      & 5:1
        & $0.328_{.007}$ & \secondcell{$0.092_{.005}$} & $0.087_{.009}$
        & $1.289_{.009}$ & $0.582_{.011}$ & $0.622_{.008}$
        & $0.403_{.006}$ & $0.293_{.011}$ & $0.308_{.009}$ \\
      & 2:1
        & $0.335_{.007}$ & $0.084_{.009}$ & $0.083_{.008}$
        & $1.425_{.010}$ & $0.555_{.009}$ & $0.610_{.010}$
        & $0.410_{.008}$ & $0.276_{.007}$ & $0.293_{.010}$ \\
      & 1:1
        & $0.329_{.011}$ & $0.086_{.004}$ & $0.081_{.007}$
        & $1.341_{.007}$ & $0.559_{.009}$ & $0.614_{.008}$
        & $0.413_{.005}$ & $0.276_{.010}$ & $0.288_{.007}$ \\
    \midrule

    \multirow{3}{*}{SRG}
      & 5:1
        & \bestcell{$0.265_{.005}$} & \bestcell{$0.107_{.004}$} & \bestcell{$0.109_{.005}$}
        & \bestcell{$0.973_{.005}$} & \bestcell{$0.639_{.006}$} & \bestcell{$0.687_{.009}$}
        & \bestcell{$0.327_{.007}$} & \bestcell{$0.312_{.005}$} & \bestcell{$0.328_{.006}$} \\
      & 2:1
        & \secondcell{$0.293_{.011}$} & $0.092_{.007}$ & \secondcell{$0.089_{.008}$}
        & \secondcell{$1.085_{.010}$} & \secondcell{$0.607_{.007}$} & \secondcell{$0.661_{.009}$}
        & \secondcell{$0.355_{.009}$} & \secondcell{$0.291_{.008}$} & \secondcell{$0.312_{.009}$} \\
      & 1:1
        & $0.322_{.006}$ & $0.081_{.007}$ & $0.076_{.003}$
        & $1.117_{.008}$ & $0.577_{.010}$ & $0.641_{.008}$
        & $0.385_{.012}$ & $0.286_{.010}$ & $0.299_{.010}$ \\
    \bottomrule
  \end{tabular}}
  \vspace{-15pt}
\end{table*}

\vspace{-4pt}
\subsection{Unseen Gene Prediction}
\label{sec:unseen}

Using PLM-derived gene semantic embeddings, TIGEM can generalize to genes absent from training. To evaluate this ability, we constructed six training–testing splits with different proportions of seen and unseen genes. The top 500 HEHVGs were used for training, while the remaining top 100, 250, and 500 HEHVGs formed three unseen test sets (ratios 5:1, 2:1, and 1:1). To further assess generalization to functionally related but unseen genes, we created another three test sets by selecting the top 100, 250, and 500 semantic-related genes (SRGs) whose PLM embeddings were closest to the training genes.

As summarized in \cref{tab:zeroshot_merged}, zero-shot performance varies across datasets. TIGEM achieves meaningful unseen-gene prediction on COAD and MEND, whereas performance on HEART is substantially weaker, indicating that zero-shot transfer remains challenging and dataset-dependent. Nevertheless, SRG test sets generally outperform their HEHVG counterparts, particularly on COAD and MEND, suggesting that PLM-derived gene semantics facilitate transfer to functionally related unseen genes. We further compare TIGEM with SGN\cite{yang2024spatial}, a recent zero-shot histology-to-ST method, under the same SRG 5:1 setting across all three datasets. TIGEM consistently outperforms SGN, with detailed results reported in \textit{Supplementary Material}.

\begin{figure}[h]
\vspace{-10pt}
  \centering
  \begin{subfigure}{0.50\linewidth}
    \centering
    \includegraphics[width=\linewidth]{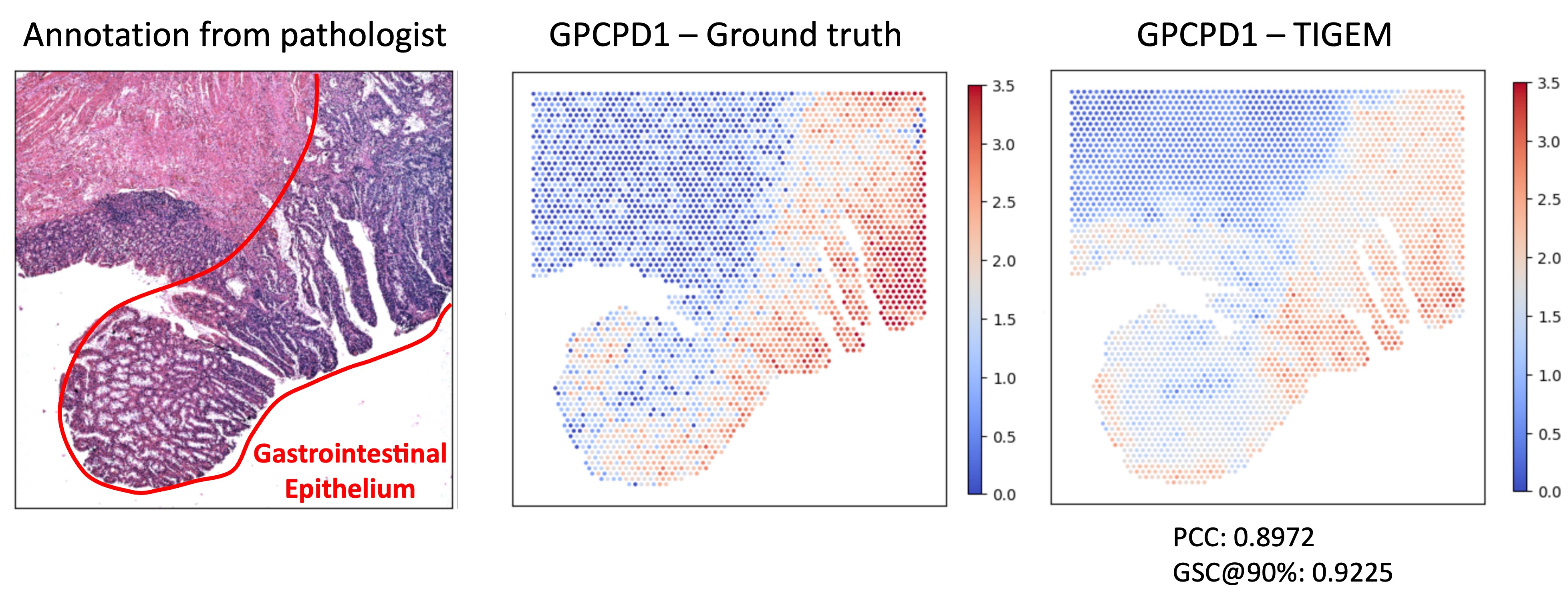}
    \caption{GPCPD1 gene prediction}
    \label{fig:short-a}
  \end{subfigure}
  \hfill
  \begin{subfigure}{0.42\linewidth}
    \centering
    \includegraphics[width=\linewidth]{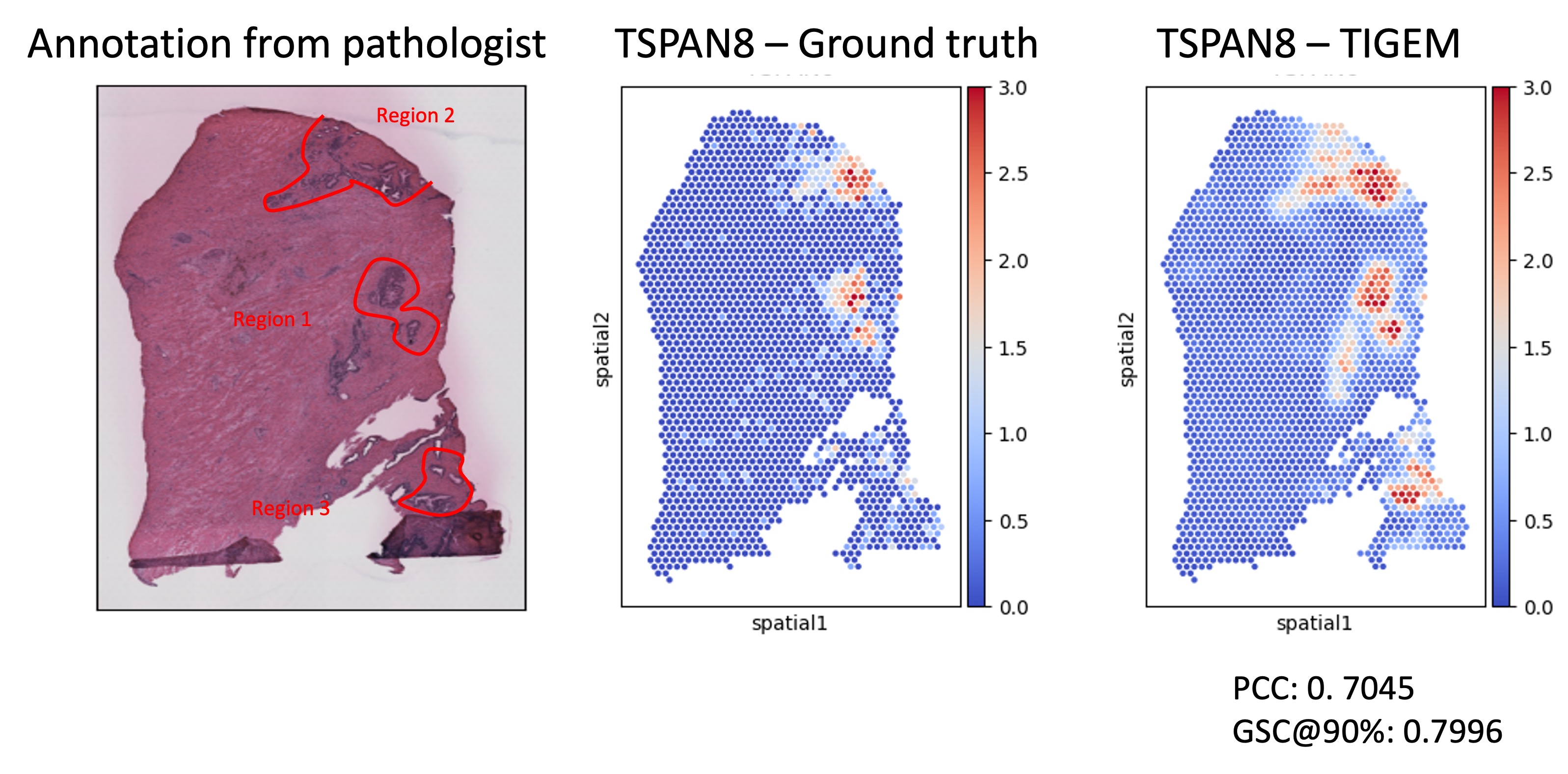}
    \caption{TSPAN8 gene prediction}
    \label{fig:short-b}
  \end{subfigure}
  \caption{Visualization of unseen biomarker gene prediction.}
  \label{fig:short}
\vspace{-10pt}
\end{figure}

Beyond aggregate performance, we further examined whether TIGEM can recover biologically meaningful spatial patterns for individual unseen genes. We evaluated the epithelial marker \textit{GPCPD1} in COAD and the prostate cancer-associated marker \textit{TSPAN8} in MEND, both excluded from training. As shown in \cref{fig:short-a}, TIGEM recovers the spatial expression pattern of \textit{GPCPD1} with high agreement with the ground truth (PCC$=0.897$, GSC@90\%$=0.922$). TIGEM also preserves the cancer-associated spatial localization of \textit{TSPAN8} in prostate tissue (\cref{fig:short-b}). These gene-level examples complement the aggregate evaluation and demonstrate that, when semantic transfer is effective, TIGEM can recover biologically relevant spatial patterns for genes unseen during training.

\vspace{-4pt}
\subsection{Extended Analysis on TIGEM}
\vspace{-4pt}
\label{sec:abl}

% ---- preamble (if not already) ----
% \usepackage[table]{xcolor}
% \newcommand{\bestcell}[1]{\cellcolor{red!20}#1}
% \newcommand{\secondcell}[1]{\cellcolor{blue!15}#1}

\begin{table*}[h]
\vspace{-10pt}
  \caption{Ablation study with LOOCV across datasets. Each variant removes a key component of TIGEM. Best and second-best results in each column are highlighted with red and blue backgrounds, respectively.}
  \label{tab:cvablation_merged_simple}
  \centering
  \small
  \renewcommand{\arraystretch}{1}
  \setlength{\tabcolsep}{4pt}

  \resizebox{\textwidth}{!}{
  \begin{tabular}{@{}c|cccc|cccc|cccc@{}}
    \toprule
    \multirow{2}{*}{Setting}
      & \multicolumn{4}{c}{HEART (LOOCV)}
      & \multicolumn{4}{c}{COAD (LOOCV)}
      & \multicolumn{4}{c}{MEND (LOOCV)} \\
    \cmidrule(lr){2-5}\cmidrule(lr){6-9}\cmidrule(lr){10-13}
      & MSE$\downarrow$ & PCC(50)$\uparrow$ & PCC(200)$\uparrow$ & GSC@90\%$\uparrow$
      & MSE$\downarrow$ & PCC(50)$\uparrow$ & PCC(200)$\uparrow$ & GSC@90\%$\uparrow$
      & MSE$\downarrow$ & PCC(50)$\uparrow$ & PCC(200)$\uparrow$ & GSC@90\%$\uparrow$ \\
    \midrule
    \grouprow{13}{Fine-grained ablation of the topology design}

    \textbf{w/o Morphology channel}
      & $0.272_{.039}$ & $0.371_{.048}$ & $0.226_{.036}$ & $0.150_{.028}$
      & $0.715_{.072}$ & $0.673_{.045}$ & $0.604_{.029}$ & $0.561_{.027}$
      & $0.325_{.037}$ & $0.685_{.048}$ & $0.539_{.038}$ & $0.414_{.047}$ 
       \\

    \textbf{w/o Topology channel}
      & $0.274_{.045}$ & $0.368_{.037}$ & $0.228_{.041}$ & $0.147_{.038}$
      & $0.720_{.064}$ & $0.676_{.061}$ & $0.599_{.048}$ & $0.553_{.034}$
      & $0.331_{.053}$ & $0.679_{.036}$ & $0.533_{.032}$ & $0.411_{.038}$ 
       \\
    \midrule
    \grouprow{13}{Broad ablation of each module}

    \textbf{w/o Residual/FFN}
      & $0.271_{.030}$ & $0.369_{.034}$ & $0.228_{.045}$ & $0.149_{.038}$
      & $0.706_{.063}$ & $0.689_{.040}$ & $0.610_{.051}$ & $0.562_{.043}$
      & $0.320_{.031}$ & $0.684_{.040}$ & $0.536_{.045}$ & $0.414_{.033}$ 
       \\

    \textbf{w/o TopoKL}
      & $0.278_{.038}$ & $0.363_{.040}$ & $0.223_{.036}$ & $0.144_{.022}$
      & $0.728_{.058}$ & $0.668_{.054}$ & $0.592_{.037}$ & $0.547_{.033}$
      & $0.329_{.028}$ & $0.679_{.063}$ & $0.531_{.035}$ & $0.409_{.034}$
\\

    \textbf{w/o GSMu}
      & \secondcell{$0.260_{.032}$} & \secondcell{$0.379_{.052}$} & \secondcell{$0.235_{.035}$} & \secondcell{$0.157_{.020}$}
      & \secondcell{$0.692_{.045}$} & \secondcell{$0.711_{.042}$} & \secondcell{$0.626_{.039}$} & \secondcell{$0.578_{.035}$}
      & \secondcell{$0.309_{.051}$} & \secondcell{$0.696_{.032}$} & \secondcell{$0.545_{.052}$} & \secondcell{$0.417_{.029}$} \\

    \midrule

    \textbf{vanilla GAT}
      & $0.284_{.037}$ & $0.358_{.044}$ & $0.219_{.053}$ & $0.144_{.022}$
      & $0.746_{.066}$ & $0.652_{.047}$ & $0.589_{.041}$ & $0.540_{.042}$
      & $0.336_{.036}$ & $0.673_{.046}$ & $0.521_{.043}$ & $0.404_{.046}$ \\

    \textbf{Full TIGEM}
      & \bestcell{$0.245_{.049}$} & \bestcell{$0.388_{.064}$} & \bestcell{$0.241_{.049}$} & \bestcell{$0.160_{.015}$}
      & \bestcell{$0.687_{.074}$} & \bestcell{$0.718_{.061}$} & \bestcell{$0.631_{.043}$} & \bestcell{$0.584_{.034}$}
      & \bestcell{$0.308_{.043}$} & \bestcell{$0.701_{.038}$} & \bestcell{$0.549_{.045}$} & \bestcell{$0.422_{.038}$} \\

    \bottomrule
  \end{tabular}}
  \vspace{-15pt}
\end{table*}

\textbf{Ablation study.}
We further perform both fine-grained and module-level ablations in \cref{tab:cvablation_merged_simple}. Full TIGEM achieves the best performance across all metrics on all three datasets. Within the topology design, removing either the morphology or topology channel consistently degrades performance, demonstrating that the two sources provide complementary cues for learning the barrier field. Removing TopoKL further reduces performance across datasets, confirming the contribution of topology-guided attention regularization. At the module level, removing Residual/FFN or GSMu also consistently degrades performance, although the effect of GSMu is relatively smaller. Finally, all TIGEM variants generally outperform vanilla GAT, supporting the benefit of the proposed topology-aware architecture over standard graph aggregation.

\begin{table}[h]
  \caption{Ablation study on the choice of PVMs (HEART dataset). Best and second-best results in each column are highlighted with red and blue backgrounds, respectively.}
  \label{tab:abl_pvm}
  \centering
  \scriptsize
  \renewcommand{\arraystretch}{1}
  \setlength{\tabcolsep}{6pt}

  \resizebox{\columnwidth}{!}{%
  \begin{tabular}{@{}lcccc@{}} \toprule
    Setting & MSE $\downarrow$ & PCC(50) $\uparrow$ & PCC(200) $\uparrow$ & GSC@90\% $\uparrow$ \\ \midrule
     CONCH only & $0.317_{.008}$ & $0.539_{.009}$ & $0.370_{.012}$ & $0.219_{.021}$ \\
     UNI-2 only & $0.306_{.002}$ & $0.545_{.005}$ & $0.378_{.010}$ & $0.227_{.013}$ \\
     GigaPath only & $0.311_{.017}$ & $0.538_{.009}$ & $0.372_{.025}$ & $0.215_{.009}$  \\ \midrule

     CONCH + UNI-2 & $0.301_{.015}$ & $0.544_{.011}$ & $0.376_{.018}$ & $0.227_{.011}$\\
     UNI-2 + GigaPath & $0.303_{.010}$ & $0.547_{.015}$ & $0.380_{.008}$ & $0.228_{.019}$ \\
     CONCH + GigaPath & $0.308_{.014}$ & $0.540_{.007}$ & $0.372_{.012}$ & $0.221_{.007}$ \\ 
     Virchow2 + UNI-2  & \secondcell{$0.291_{.016}$} & \bestcell{$0.552_{.010}$} & $0.381_{.024}$ & \bestcell{$0.230_{.011}$} \\ \midrule

     CONCH + UNI-2 + GigaPath & $0.294_{.016}$ & $0.548_{.017}$ & $0.377_{.018}$ & $0.226_{.011}$ \\
     CONCH + Virchow2 + GigaPath & $0.292_{.011}$ & $0.547_{.009}$ & $0.381_{.006}$ & \secondcell{$0.229_{.008}$} \\ 
     Virchow2 + UNI-2 + GigaPath & $0.291_{.012}$ & \secondcell{$0.551_{.016}$} & \secondcell{$0.382_{.008}$} & $0.226_{.019}$ \\
     CONCH + UNI-2 + GigaPath + Virchow2 & \bestcell{$0.289_{.017}$} & $0.550_{.025}$ & \bestcell{$0.383_{.018}$} & \bestcell{$0.230_{.014}$} \\ \bottomrule
  \end{tabular}}
  \vspace{-10pt}
\end{table}

\textbf{Choice of PVM encoders.}
We benchmark several pretrained PVM backbones and combinations~\cite{vorontsov2024virchow, lu2024avisionlanguage, chen2024uni, xu2024whole}. As shown in \cref{tab:abl_pvm}, multi-PVM combinations consistently outperform single backbones, indicating complementary feature scales. Among two-model pairings, Virchow2+UNI-2 offers the best overall balance, reaching top GSC@90\% with competitive MSE/PCC. Adding more PFMs yields only marginal gains (e.g., slightly better MSE/PCC(200)) while substantially increasing preprocessing cost, so we adopt a compact but effective combination in our main experiments.

\textbf{Zero-shot evidence.}
To justify the feasibility of strict $G_{\mathrm{test}}\cap G_{\mathrm{train}}=\emptyset$ inference with gene embedding, we added sensitivity analyses (\cref{tab:fe}) on MEND and COAD under the SRG 5:1 split and compared TIGEM with non-semantic conditioning baselines, including random gene embeddings and one-hot gene identifiers. Semantic gene/protein embedding variants substantially outperform these baselines, indicating that the gain does not come from simply providing a gene identifier but from meaningful gene-level semantics. The additional dataset is presented in \textit{Supplementary Material}.

\begin{table}[h]
  \centering
  \scriptsize
  \vspace{-4pt}
  \caption{Justification of the feasibility of strict $G_{\mathrm{test}}\cap G_{\mathrm{train}}=\emptyset$ inference.}
  \renewcommand{\arraystretch}{1}
  \setlength{\tabcolsep}{1pt}
  \resizebox{\columnwidth}{!}{%
  \begin{tabular}{@{}lccc ccc@{}}
    \toprule
    \multirow{2}{*}{Setting} 
    & \multicolumn{3}{c}{COAD} 
    & \multicolumn{3}{c}{MEND} \\
    \cmidrule(lr){2-4} \cmidrule(lr){5-7}
    & MSE $\downarrow$ 
    & PCC (All) $\uparrow$ 
    & GSC@90\% $\uparrow$
    & MSE $\downarrow$ 
    & PCC (All) $\uparrow$ 
    & GSC@90\% $\uparrow$ \\ 
    \midrule
    SRG 5:1 
    & \bestcell{$0.973_{.005}$} & \bestcell{$0.639_{.006}$} & \bestcell{$0.687_{.009}$}
    & \bestcell{$0.327_{.007}$} & \bestcell{$0.312_{.005}$} & \bestcell{$0.328_{.006}$} \\
    Random 
    & \secondcell{$1.585_{.010}$} & $0.422_{.009}$ & $0.459_{.010}$
    & $0.450_{.006}$ & $0.199_{.006}$ & \secondcell{$0.192_{.007}$} \\
    One-hot 
    & $1.596_{.009}$ & \secondcell{$0.425_{.008}$} & \secondcell{$0.464_{.005}$}
    & \secondcell{$0.431_{.008}$} & \secondcell{$0.203_{.005}$} & $0.185_{.006}$ \\
    \bottomrule
  \end{tabular}
  }
  \label{tab:fe}
  \vspace{-5pt}
\end{table}

\textbf{Effect of hyperparameters and model scale.}
We next analyzed the impact of key hyperparameters, including the temperature $\tau$, the KL weight $\lambda_{topo}$, and the number of GAT layers $L$ in the TIG module. The result figures provided in the \textit{Supplementary Material} show that lowering the temperature (sharper topology prior) or increasing $\lambda_{topo}$ consistently improves model performance, supporting the importance of the topology prior for guiding spatial prediction.

We also assessed model scaling by varying the number of GAT layers in the TIG module. As reported in the \textit{Supplementary Material}, moderate increases in depth improve accuracy, but performance plateaus once the model reaches 3–5 blocks, indicating limited benefit from further growth.

\textbf{Computational Cost.}
We further summarized the computational cost of TIGEM and all baselines on a single-slide sample per LOOCV fold, with all runs conducted on the same NVIDIA A40 GPU. The results are provided in the \textit{Supplementary Material}. We report parameter counts and wall-clock training time (to convergence/early stopping) for all methods, while inference time is reported only for Stem since its DiT-based sampling introduces substantial prediction overhead. Notably, TIGEM’s training time already includes patch embedding extraction; with pre-generated patch embeddings and cached topology priors, TIGEM avoids training image encoders from scratch and is consistently among the fastest methods. In contrast, Stem is orders of magnitude slower.

\vspace{-7pt}
\section{Conclusion}
\vspace{-2pt}
We present \textbf{TIGEM}, a topology-informed graph attention framework with gene semantic modulation for histology-to-spatial transcriptomics prediction. TIGEM regularizes graph message passing with a topology-guided prior distribution to reduce cross-region over-smoothing, and decodes expression through a gene-conditioned GSMu head, enabling strict zero-shot inference for unseen genes by swapping in gene embeddings at test time. We also introduce \textbf{GSC}, a complementary graph-spectral metric for evaluating spatial-pattern fidelity beyond conventional spot-wise metrics. Across diverse ST datasets, TIGEM consistently outperforms existing methods under MSE, PCC, and GSC, while improvements in the established SSIM metric independently corroborate the spatial fidelity captured by GSC. Ablation studies further demonstrate consistent contributions from the topology barrier field, TopoKL, and GSMu, while zero-shot experiments show the potential to recover biologically relevant expression patterns for genes absent from training. Together, these results establish TIGEM as a generalizable framework that integrates topology-aware spatial modeling with gene-semantic conditioning for histology-to-ST prediction.

{
    \small
    \bibliographystyle{ieeenat_fullname}
    \bibliography{main}
}
\clearpage
\setcounter{page}{1}
\maketitlesupplementary

\section{Dataset Description}
\label{sec:data}

In this work, we benchmarked our method on three large-scale ST datasets, whose details are summarized in \cref{tab:dt}.

\begin{table}[h]
  \caption{Datasets statistics}
  \label{tab:dt}
  \centering
  \small
  \renewcommand{\arraystretch}{0.9}
  \setlength{\tabcolsep}{10pt}

  \resizebox{\columnwidth}{!}{%
  \begin{tabular}{@{}cccccc@{}} \toprule
    Dataset & Technology & \#Samples & Avg. spots & Total spots & Organ\\ \midrule
      HEART & 10X Visium & 9 & 2461 & 25861 & Heart \\
      COAD & 10X Visium & 13 & 3354 & 43605 & Bowel\\
      MEND & 10X Visium & 14 & 2455 & 34370 & Prostate \\ \bottomrule
  \end{tabular}}
  \vspace{-0.6em}
\end{table}

All three datasets were sourced from HEST-1K\cite{jaume2024hest}. The HEART dataset contains samples MISC132–142 (excluding MISC137, 134), the COAD dataset includes samples MISC33–45, and the MEND dataset contains samples 140–153. Raw expression counts were transformed by the log1p numpy function. All experiment results can be reproduced by directly downloading the datasets we mentioned without further preprocessing. 

\section{Topology cue construction}
\label{sec:supp_topo_cues}

We construct three topology cue channels from CellViT cell segmentation~\cite{horst2024cellvit} to form
$F_{\mathrm{topo}}=[F_{\mathrm{density}}\;\|\;F_{\mathrm{count}}\;\|\;F_{\mathrm{anisotropy}}]$.
CellViT provides a set of nuclei (or cell) centroids
$\mathcal{C}=\{\mathbf{c}_m=(x_m,y_m)\}_{m=1}^{M}$ in the same slide coordinate system as spot centers.
All topology maps are computed on a regular grid of size $H_f\times W_f$ aligned with the WSI,
with grid location $\mathbf{u}$ representing the corresponding pixel coordinate.

\paragraph{Kernelized density map.}
We define a smooth cell density field by Gaussian kernel density estimation (KDE):
\begin{equation}
F_{\mathrm{density}}(\mathbf{u})
=
\sum_{m=1}^{M}
K_{\sigma_d}\!\left(\mathbf{u}-\mathbf{c}_m\right),
\qquad
K_{\sigma}(\mathbf{r})=\exp\!\left(-\frac{\|\mathbf{r}\|_2^2}{2\sigma^2}\right),
\label{eq:density_kde}
\end{equation}
where $\sigma_d$ controls the spatial smoothing bandwidth.
This channel captures local cellularity and highlights tissue compartments with distinct cell packing.

\paragraph{Local nuclei count map.}
To provide a complementary, less-smoothed signal, we compute a local count within a fixed-radius neighborhood:
\begin{equation}
F_{\mathrm{count}}(\mathbf{u})
=
\sum_{m=1}^{M}
\mathbf{1}\!\left[\;\|\mathbf{u}-\mathbf{c}_m\|_2 \le r_c\;\right],
\label{eq:count}
\end{equation}
where $r_c$ is a counting radius (in pixels).
Compared with KDE density, this produces a piecewise-constant map that is sensitive to sharp transitions.

\paragraph{Anisotropy / structure map.}
We quantify local directional organization by measuring the anisotropy of the centroid distribution in a neighborhood.
For each grid point $\mathbf{u}$, let $\mathcal{N}(\mathbf{u})=\{m:\|\mathbf{u}-\mathbf{c}_m\|_2\le r_a\}$ be nearby centroids.
We compute the weighted second-moment matrix
\begin{equation}
S(\mathbf{u})
=
\frac{1}{Z(\mathbf{u})}
\sum_{m\in\mathcal{N}(\mathbf{u})}
w_m(\mathbf{u})
\left(\mathbf{c}_m-\boldsymbol{\mu}(\mathbf{u})\right)
\left(\mathbf{c}_m-\boldsymbol{\mu}(\mathbf{u})\right)^{\!\top},
\label{eq:moment}
\end{equation}
where $w_m(\mathbf{u})=K_{\sigma_a}(\mathbf{u}-\mathbf{c}_m)$,
$Z(\mathbf{u})=\sum_{m\in\mathcal{N}(\mathbf{u})} w_m(\mathbf{u})$,
and
$\boldsymbol{\mu}(\mathbf{u})=\frac{1}{Z(\mathbf{u})}\\\sum_{m\in\mathcal{N}(\mathbf{u})} w_m(\mathbf{u})\,\mathbf{c}_m$.
Let $\lambda_1(\mathbf{u})\ge \lambda_2(\mathbf{u})\ge 0$ denote eigenvalues of $S(\mathbf{u})$.
We define an anisotropy index
\begin{equation}
F_{\mathrm{anisotropy}}(\mathbf{u})
=
\frac{\lambda_1(\mathbf{u})-\lambda_2(\mathbf{u})}{\lambda_1(\mathbf{u})+\lambda_2(\mathbf{u})+\varepsilon},
\label{eq:anisotropy}
\end{equation}
where $\varepsilon$ is a small constant.
This channel is high in locally aligned or elongated structures (e.g., stromal bands) and low in isotropic regions,
providing cues about boundaries and compartment transitions.

\paragraph{Normalization and stacking.}
Before concatenation, each channel is robustly normalized per slide:
\begin{equation}
\tilde F(\mathbf{u})=\mathrm{clip}\!\left(\frac{F(\mathbf{u})-\mathrm{median}(F)}{\mathrm{IQR}(F)+\varepsilon},\,[-3,3]\right),
\label{eq:robust_norm}
\end{equation}
and we set $F_{\mathrm{topo}}=[\tilde F_{\mathrm{density}}\;\|\;\tilde F_{\mathrm{count}}\;\|\;\tilde F_{\mathrm{anisotropy}}]$.

In summary, these three channels are chosen to provide a compact but expressive summary of cell-level geometry that is most relevant to spatial expression inference.
$F_{\mathrm{density}}$ captures \emph{cellularity} and coarse compartmental structure (e.g., tumor-rich vs.\ stroma-rich regions), which is strongly associated with broad transcriptional programs.
$F_{\mathrm{count}}$ complements KDE density by retaining \emph{sharp local discontinuities} and abrupt changes in cellular abundance that frequently occur near lumens, invasive fronts, and tissue interfaces—precisely where message passing should be attenuated.
$F_{\mathrm{anisotropy}}$ encodes \emph{organization and alignment} of local cell arrangements, highlighting directional stromal bands, glandular boundaries, and other architectural motifs that are not well represented by scalar density alone.
Together, these cues cover three fundamental geometric factors—\emph{how many cells are present} (mass), \emph{how abruptly this changes} (discontinuity), and \emph{how cells are arranged} (orientation/structure)—which empirically explain most topology-driven variation in tissue microenvironments at Visium resolution.
We therefore treat $F_{\mathrm{topo}}=[F_{\mathrm{density}}\|\;F_{\mathrm{count}}\|\;F_{\mathrm{anisotropy}}]$ as a lightweight yet sufficient topology descriptor set for learning slide-specific barrier fields, while keeping the field branch inexpensive and robust to segmentation noise.

\subsection{Constructing positive/negative edge sets for barrier-field supervision}
\label{supp:posneg_edges}

We build a spot graph $\mathcal{G}=(\mathcal{V},\mathcal{E})$ (Visium 6-neighbors MNN). For each edge $(i,j)\in\mathcal{E}$, we compute two scores:

\textbf{Morphology discrepancy.}
\begin{equation}
d^{\mathrm{morph}}_{ij} = \|\mathbf x_i-\mathbf x_j\|_2,
\label{eq:supp_morph_dist}
\end{equation}
where $\mathbf x_i$ is the patch embedding.

\textbf{Topology discrepancy.}
Let $\mathbf t_i\in\mathbb{R}^{3}$ be the topology descriptor at spot $i$, obtained by sampling
$[F_{\mathrm{density}},F_{\mathrm{count}},F_{\mathrm{anis}}]$ at $\mathbf s_i$.
\begin{equation}
d^{\mathrm{topo}}_{ij} = \|\mathbf t_i-\mathbf t_j\|_2.
\label{eq:supp_topo_dist}
\end{equation}

We define positive edges $\mathcal P$ (likely within-compartment) and negative edges $\mathcal N$ (likely cross-boundary) by thresholding these discrepancies using per-slide quantiles:
\begin{align}
(i,j)\in\mathcal P &\iff d^{\mathrm{morph}}_{ij} \le Q_{\alpha}\!\left(d^{\mathrm{morph}}\right)\ \land\
d^{\mathrm{topo}}_{ij} \le Q_{\alpha}\!\left(d^{\mathrm{topo}}\right), \label{eq:supp_pos}\\
(i,j)\in\mathcal N &\iff d^{\mathrm{morph}}_{ij} \ge Q_{\beta}\!\left(d^{\mathrm{morph}}\right)\ \lor\
d^{\mathrm{topo}}_{ij} \ge Q_{\beta}\!\left(d^{\mathrm{topo}}\right), \label{eq:supp_neg}
\end{align}
with $\alpha<\beta$ (e.g., $\alpha=0.3,\beta=0.7$). This yields weak but effective pseudo-labels for boundary supervision.

\paragraph{Barrier-field loss.}
Given the barrier map $B=f_{\mathrm{field}}(F)$, we compute the traversal cost $c_{ij}$ along the straight segment between $\mathbf s_i$ and $\mathbf s_j$ (Eq.~10 in the main text).
We supervise $B$ by enforcing that positive edges have lower costs than negative edges incident to the same anchor node $i$:
\begin{equation}
\mathcal L_{\mathrm{field}}
=
\frac{1}{|\mathcal P|}
\sum_{(i,j)\in\mathcal P}
\frac{1}{|\mathcal N(i)|}
\sum_{(i,k)\in\mathcal N(i)}
\max\!\big(0,\; m + c_{ij} - c_{ik}\big),
\label{eq:supp_loss_field}
\end{equation}
where $\mathcal N(i)=\{(i,k)\in\mathcal N\}$ and $m>0$ is a margin.
This provides a direct training signal for $f_{\mathrm{field}}$ and avoids degeneracy of the barrier map.

The barrier field $B$ is optimized to \emph{rank} traversal costs on local graph edges: for an anchor spot $i$, edges in $\mathcal P$ (likely within the same compartment/state) are enforced to have lower line-integral cost $c_{ij}$ than edges in $\mathcal N(i)$ (likely crossing a boundary), up to a margin $m$ (Eq.~\eqref{eq:supp_loss_field}). Since $c_{ij}$ aggregates $B$ along the straight segment connecting $\mathbf s_i$ and $\mathbf s_j$, the only way to satisfy these inequalities consistently across many anchors is to assign high $B$ values to spatial locations that are repeatedly traversed by negative edges but rarely by positive edges. These locations correspond to \emph{persistent transition regions} in the tissue plane (e.g., lumens, stromal interfaces, compartment borders), while homogeneous regions accumulate low $B$. 

As a result, $B$ induces a \emph{data-driven propagation prior} on the spot graph: edges whose connecting paths pass through high-barrier regions receive large $c_{ij}$ and are down-weighted, whereas edges staying within low-barrier regions receive small $c_{ij}$ and are favored. Converting costs to a neighborhood distribution via
$q_{ij}\propto \exp(-c_{ij}/\tau_b)$ therefore yields a topology-aware teacher graph that preserves within-compartment connectivity while suppressing cross-boundary leakage. Importantly, this supervision is independent of the GAT attention and does not rely on matching the student; thus the learned field does not trivially collapse to the model's current attention pattern and provides a stable geometric prior for boundary-aware message passing.

\section{Pseudocode of TIGEM}

Due to restrictions on external links, we provide pseudocode for our implementation in the manuscript instead of sharing an anonymous GitHub repository. Following the modular structure of our codebase, we decompose the framework into four components and present pseudocode for each. The topology prior construction is given in \cref{alg:build-graph-topo}, train barrier field is shown in \cref{alg:train_bf} and TIG block details are shown in \cref{alg:tig-kl-combined}, the GSMu algorithm is provided in \cref{alg:gene-head}, and unseen gene prediction is provided in \cref{alg:zero-shot}. The official coda base will be provided after being accepted.

\begin{algorithm}[h]
\caption{Topology prior construction for TIGEM}
\label{alg:build-graph-topo}
\begin{algorithmic}[1]
\Require Spot coordinates $\{\mathbf s_i\}_{i=1}^N$, patch embeddings $\{\mathbf x_i\}_{i=1}^N$, CellViT cell centroids $\mathcal C=\{(u_m,v_m)\}_{m=1}^M$
\Require Hyperparameters: $k$ (neighbors), grid size $(H_f,W_f)$, splat bandwidth $\sigma$, thresholds $(\tau_{\text{dens}},\tau_{\text{ani}},\tau_{\text{morph}})$
\Ensure Spot graph $\mathcal G=(\mathcal V,\mathcal E)$, positive edges $\mathcal P$, negative edges $\mathcal N$

\State \textbf{Build spot graph} $\mathcal E \leftarrow$ MNN($\{\mathbf s_i\}$, $k$) \Comment{or Visium 6-neighbor}
\State \textbf{Rasterize morphology} $F_{\text{morph}} \leftarrow \sum_{i=1}^N K_\sigma(\cdot-\mathbf s_i)\, W_x\mathbf x_i$
\State \textbf{Compute topology cues from }$\mathcal C$:
\State \quad density map $F_{\text{dens}}$; anisotropy map $F_{\text{ani}}$; nuclei-count map $F_{\text{cnt}}$
\State $F_{\text{topo}} \leftarrow [F_{\text{dens}} \,\|\, F_{\text{cnt}} \,\|\, F_{\text{ani}}]$
\State $F \leftarrow [F_{\text{morph}} \,\|\, F_{\text{topo}}] \in \mathbb R^{C\times H_f\times W_f}$

\State \textbf{Score each edge} $(i,j)\in\mathcal E$ using cue differences:
\State \quad $d^{\text{morph}}_{ij}\leftarrow \|W_m\mathbf x_i - W_m\mathbf x_j\|_2$
\State \quad $d^{\text{dens}}_{ij}\leftarrow |F_{\text{dens}}(\mathbf s_i)-F_{\text{dens}}(\mathbf s_j)|$
\State \quad $d^{\text{ani}}_{ij}\leftarrow |F_{\text{ani}}(\mathbf s_i)-F_{\text{ani}}(\mathbf s_j)|$
\State \quad $r_{ij}\leftarrow \mathds{1}[d^{\text{morph}}_{ij}<\tau_{\text{morph}}]\cdot\mathds{1}[d^{\text{dens}}_{ij}<\tau_{\text{dens}}]\cdot\mathds{1}[d^{\text{ani}}_{ij}<\tau_{\text{ani}}]$

\State \textbf{Construct pseudo-label sets}
\State $\mathcal P \leftarrow \{(i,j)\in\mathcal E: r_{ij}=1\}$ \Comment{likely within-compartment}
\State $\mathcal N \leftarrow \{(i,j)\in\mathcal E: r_{ij}=0\}$ \Comment{likely cross-boundary}
\State \Return $\mathcal G,\,\mathcal P,\,\mathcal N,\,F$
\end{algorithmic}
\end{algorithm}

\begin{algorithm}[h!]
\caption{Training the barrier field branch}
\label{alg:train_bf}
\begin{algorithmic}[1]
\Require Fused feature map $F$, spot coords $\{\mathbf s_i\}$, edge sets $\mathcal P,\mathcal N$, temperature $\tau_b$, margin $m$, samples $T$
\Require Barrier network $f_{\text{field}}(\cdot)$, optimizer \textsc{Opt}, training steps $S$
\Ensure Trained barrier field network $f_{\text{field}}$

\For{$t=1$ to $S$}
  \State $B \leftarrow \mathrm{softplus}(f_{\text{field}}(F))$ \Comment{nonnegative barrier map}
  \State $\mathcal L_{\text{field}}\leftarrow 0$
  \ForAll{$(i,j)\in\mathcal P$}
    \State Sample negatives $\mathcal N(i)=\{(i,k)\in\mathcal N\}$ (or a subset)
    \State $c_{ij}\leftarrow \mathrm{LineCost}(B,\mathbf s_i,\mathbf s_j,T)$
    \ForAll{$(i,k)\in\mathcal N(i)$}
      \State $c_{ik}\leftarrow \mathrm{LineCost}(B,\mathbf s_i,\mathbf s_k,T)$
      \State $\mathcal L_{\text{field}} \mathrel{+}= \max(0,\, m + c_{ij}-c_{ik})$
    \EndFor
  \EndFor
  \State $\mathcal L_{\text{field}}\leftarrow \mathcal L_{\text{field}} / (|\mathcal P|\cdot \mathbb E|\mathcal N(i)|)$
  \State \textsc{Opt}.zero\_grad(); \; backprop $\nabla \mathcal L_{\text{field}}$; \; \textsc{Opt}.step()
\EndFor
\State \Return $f_{\text{field}}$
\end{algorithmic}
\end{algorithm}

\begin{algorithm}[t]
\caption{\textsc{LineCost}$(B,\mathbf s_i,\mathbf s_j,T)$}
\label{alg:linecost}
\begin{algorithmic}[1]
\Require Barrier map $B$, endpoints $\mathbf s_i,\mathbf s_j$, samples $T$
\Ensure Traversal cost $c_{ij}$
\State $\ell_{ij}(t)\leftarrow (1-t)\mathbf s_i+t\mathbf s_j$
\State $\{\tau_r\}_{r=1}^T \leftarrow$ uniform grid in $[0,1]$
\State $c_{ij}\leftarrow \frac{\|\mathbf s_i-\mathbf s_j\|_2}{T}\sum_{r=1}^T B(\ell_{ij}(\tau_r))$
\State \Return $c_{ij}$
\end{algorithmic}
\end{algorithm}

\begin{algorithm}[h!]
\caption{TIG Block with Barrier-Field Teacher and Topology-KL}
\label{alg:tig-kl-combined}
\begin{algorithmic}[1]

\Require \parbox[t]{0.92\linewidth}{
Prev features $H^{(\ell-1)}\in\mathbb{R}^{N\times d}$; spot graph $edge\_index$; spot coordinates $\{ \mathbf{s}_i\}_{i=1}^{N}$; 
fused raster map $F$; barrier network $f_{\mathrm{field}}$; temperature $\tau_b$; samples $T$; 
TopoKL weight $\lambda_{\mathrm{topo}}$; task loss $\mathcal{L}_{\mathrm{task}}$.
}
\Ensure \parbox[t]{0.92\linewidth}{
Updated features $H^{(\ell)}$ and total loss $\mathcal{L}$.
}

\State \textbf{(Barrier-field teacher)} \;
$B \gets \mathrm{softplus}\!\big(f_{\mathrm{field}}(F)\big)$ \Comment{$B\in\mathbb{R}^{H_f\times W_f}$, nonnegative}
\Statex \hspace{\algorithmicindent}\parbox[t]{0.92\linewidth}{
For each edge $(i,j)\in edge\_index$, compute line cost
$c_{ij}\gets \mathrm{LineCost}(B,\mathbf{s}_i,\mathbf{s}_j,T)$ by sampling $T$ points on the segment.
}
\Statex \hspace{\algorithmicindent}\parbox[t]{0.92\linewidth}{
Teacher distribution: $q_{ij}\gets \mathrm{softmax}_{j\in\mathcal{N}(i)}\!\left(-c_{ij}/\tau_b\right)$,
and we detach it: $\tilde q \gets \mathrm{stopgrad}(q)$.
}
\Statex

\State \textbf{(Student attention + TIG update)} \;
$Z \gets \mathrm{LN}\!\big(H^{(\ell-1)}\big)$
\State $(Z_{\text{gat}},\,(eidx,\alpha)) \gets \mathrm{MultiHead\text{-}GATv2}(Z,\;edge\_index)$ \Comment{$\alpha\in\mathbb{R}^{E\times M}$}
\State $Z_{\text{gat}} \gets \mathrm{Dropout}\!\left(\mathrm{ELU}(Z_{\text{gat}})\right)$
\State $\tilde{H} \gets \mathrm{Proj}\!\left(H^{(\ell-1)}\right) + Z_{\text{gat}}$ \Comment{Proj only if dims differ}
\State $H^{(\ell)} \gets \tilde{H} + \mathrm{FFN}\!\big(\mathrm{LN}(\tilde{H})\big)$
\Statex

\State \textbf{(Topology-KL regularization)}
\Statex \hspace{\algorithmicindent}\parbox[t]{0.92\linewidth}{
Mask non-self: $M\gets\{k~|~eidx_0[k]\neq eidx_1[k]\}$; $src\gets eidx_0[M]$, $\alpha\gets \alpha[M]$.
}
\State $p_{\text{edge}}\gets \mathrm{mean\_heads}(\alpha)$
\State $p \gets \mathrm{softmax\_by\_src}(p_{\text{edge}},\,src)$ \Comment{student probs aligned with $(src,dst)$}
\Statex \hspace{\algorithmicindent}\parbox[t]{0.92\linewidth}{
Align $\tilde q$ to current $(src,dst)$ order (same edge list); obtain $\tilde q$ per source node.
}
\State $\displaystyle \mathcal{L}^{(\ell)}_{\text{topo}} \gets \frac{1}{|B|}\sum_{i}\sum_{j\in\mathcal{N}(i)} 
\tilde q_{ij}\left(\log(\tilde q_{ij}+\epsilon)-\log(p_{ij}+\epsilon)\right)$
\State $\mathcal{L} \gets \mathcal{L}_{\mathrm{task}} + \lambda_{\mathrm{topo}}\mathcal{L}^{(\ell)}_{\text{topo}}$
\State \Return $H^{(\ell)}$, $\mathcal{L}$

\end{algorithmic}
\end{algorithm}

\begin{algorithm}[h!]
\caption{GSMu}
\label{alg:gene-head}
\begin{algorithmic}[1]
\Require Node embeddings $H\in\mathbb{R}^{N\times d}$; gene embeddings $E\in\mathbb{R}^{G\times d_g}$; projection $W_e$
\Ensure Predictions $\hat{Y}\in\mathbb{R}^{N\times G}$
\State $\Phi \gets \mathrm{LN}(E)\, W_e \in \mathbb{R}^{G\times d}$ \Comment{align genes to encoder dim}
\State \textbf{(Cross-attention)} For each gene $g$: $\; Z_g \gets \mathrm{CrossAttn}(q=\Phi_g, K=H, V=H)\in\mathbb{R}^{d}$
\State \textbf{(FiLM)} $(\gamma_g,\beta_g) \gets \mathrm{MLP}_{\text{FiLM}}(Z_g)$ with $\gamma_g\approx 1, \beta_g\approx 0$ at init (tanh-gated)
\State Shared readout $u\in\mathbb{R}^{d}$
\State $\hat{Y}_{i,g} \gets u^\top \big( \gamma_g \odot H_i + \beta_g \big)$ \quad (apply Softplus to ensure non-negativity)
\State \Return $\hat{Y}$
\end{algorithmic}
\end{algorithm}

\begin{algorithm}
\caption{Zero-Shot Gene Prediction}
\label{alg:zero-shot}
\begin{algorithmic}[1]
\Require Unseen gene embedding $e_{g^\ast}\in\mathbb{R}^{d_g}$; frozen encoder output $H$
\Ensure $\hat{y}_{:,g^\ast}\in\mathbb{R}^{N}$
\State $\Phi_{g^\ast} \gets \mathrm{LN}(e_{g^\ast})\, W_e$
\State $Z_{g^\ast} \gets \mathrm{CrossAttn}(q=\Phi_{g^\ast}, K=H, V=H)$
\State $(\gamma_{g^\ast},\beta_{g^\ast}) \gets \mathrm{MLP}_{\text{FiLM}}(Z_{g^\ast})$
\State $\hat{y}_{i,g^\ast} \gets u^\top \left( \gamma_{g^\ast} \odot H_i + \beta_{g^\ast} \right)$
\State \Return $\hat{y}_{:,g^\ast}$
\end{algorithmic}
\end{algorithm}

\section{TIGEM implementation and training}

Here we provide the implementation details of our TIGEM for reproduction. Generally, our model is implemented based on Pytorch v2.7.

\textbf{Backbone encoder.}
We stack $L{=}3$ identical \emph{TIG} blocks. Each block applies Pre-Norm LayerNorm, a multi-head GATv2 layer on the spot graph ($H{=}4$ heads, hidden $d{=}512$), ELU and dropout ($p{=}0.2$), followed by a residual projection and a position-wise FFN (GELU, expansion ratio $2\times$) with a second residual. The topology appears via a KL regularizer.

\textbf{Node features and graphs.}
For each slide we concatenate UNI-2 and Virchow2 embeddings. The spot graph is built by kNN$\rightarrow$MNN with $k{=}6$ on full-resolution pixel coordinates and pruned by a global threshold $\tau{=}c\cdot\text{median-NN}$ with $c{=}1.5$; self-loops are removed and the graph is symmetrized.

\textbf{Training.}
We optimize the mean squared error on $\log(1{+}\text{counts})$ plus $\lambda_{\text{topo}}\mathcal{L}_{\text{topo}}$ with AdamW (lr $10^{-4}$, weight decay $10^{-5}$), gradient clipping (5.0), batch size $2$ slides, $500$ epochs.
Weights are Xavier-initialized; we fix random seeds and use deterministic cuDNN settings.

\textbf{Inference and zero-shot.}
Given a trained encoder, unseen genes $g^\ast$ are handled without retraining: compute the ESM2 embedding, project to a gene token, run the cross-attention$+$FiLM head, and read out $\hat{Y}_{\cdot,g^\ast}$.
The topology prior is reused as is.

\section{Experiment Implementation}
\label{sec:rationale}

\paragraph{Running environment}
The experiments and model training are conducted on two Linux servers. Server 1: Most of the experiments are conducted on the server with the AMD EPYC 7452 32-Core Processor, 503G RAM, and 4 Nvidia A40 48 GB GPUs. Server 2: To speed up the model training, we put the heavy tasks (including Stem model training and inference and TIGEM training) on the server with the AMD EPYC 9354 32-Core Processor, 1.0T RAM, and 1 Nvidia H200 141GB GPU. 

\paragraph{Baseline training.} This subsection presents the implementation details of the baseline models to guarantee a consistent and fair comparison with TIGEM. All input preprocessing follows the procedures described in \cref{sec:data}.

\textbf{ST-Net.} ST-Net adopts a DenseNet pretrained on ImageNet as its initialization and fine-tunes it on ST data by substituting the final regression layer. In our implementation, we set the learning rate as $10^{-4}$ while maintaining other hyperparameters and the architectural design reported in the original work. %其中，epoch是多少？batch size是？

\textbf{HistoGene.} This model employs a Vision Transformer (ViT) to encode each spot-level image patch from the WSI. Following the configuration reported in the original work, we use 4 transformer layers with 16 attention heads, a hidden dimension of 128, and a dropout rate of~0.1. 
Spot coordinates are projected through a linear embedding layer before being fused with visual features. 
% During training, spot patches undergo random horizontal flipping and rotation, followed by normalization using the ImageNet statistics. 

\textbf{THItoGene.}
THItoGene extracts molecular features from each spot’s patch using a dynamic convolution module and an Efficient-Capsule block. 
Spot coordinates are embedded through a positional encoding module. 
These visual and positional features are then globally modeled with a ViT to capture long-range spatial relationships. 
Finally, a GAT layer adaptively aggregates neighborhood information to predict gene expression. Similar to HistoGene, we use 4 transformer layers. Learning rate is set as 1e-5. % attn layer的数目是4， 且学习率也是1e-5, attn head他们设置了两个不同的，有点麻烦就不说了

\textbf{BLEEP.}
BLEEP learns a joint embedding space for histology images and gene expression profiles through a contrastive objective.  For each spot, the model retrieves the $k$ most similar spots from a reference pool and uses their averaged expression as the prediction. In our implementation, a pretrained pathology foundation model serves as the visual encoder, while shallow MLP heads project both image features and gene vectors into the shared space. We used batch size 256 and learning rate $5\times10^{-4}$. We set the contrastive temperature to~1 and use $k=50$ retrieved neighbors.

\textbf{TRIPLEX.}
TRIPLEX integrates information from multiple spatial scales through four modules: a target encoder, a local neighborhood encoder, a global WSI encoder, and a fusion encoder that aggregates their outputs. 
Following the official setup, each spot image is first embedded with a pretrained pathology foundation model before entering the network. According to the original paper, the APEG module is used in the model. Default settings are used in the benchmarking.
% All encoder blocks use 2 layers, 8 attention heads, and a dropout rate of 0.15, and the local module incorporates features from the 25 nearest spots.  Spot coordinates are encoded using the convolutional positional encoding module proposed in the original work.   % 他们的默认设置太多，检查不完也描述不完，我感觉就说APEG模块的就行吧，另外我们所有模型没用图像增强

\textbf{Stem.}
Stem adopts a Diffusion Transformer (DiT)–based conditional generative model to predict spot-level gene expression. During training, each patch is encoded by a pretrained pathology foundation model, and the resulting embedding is used as the conditioning signal for denoising the gene expression vector. 
At inference time, Stem generates multiple samples and averages them to obtain the final prediction. 
We used a batch size of 128 and $5\times10^{-5}$ learning rate with 300K iterations for all experiments.

\textbf{M2TGLGO.} M2TGLGO models gene relationships and spatial information for histology-based spatial gene expression prediction. In our implementation, we follow the architectural design and training strategy reported in the original work and evaluate the model using the same data splits as TIGEM.

\textbf{STFlow.} STFlow employs a flow-based generative framework to predict spatial gene expression from histology images. In our implementation, we follow the network architecture and hyperparameter settings reported in the original work and retrain the model using the same training and testing splits as TIGEM.

\textbf{MERGE.} MERGE integrates histological representations and spatial information for spatial gene expression prediction. In our implementation, we follow the original model architecture and training configuration and evaluate it under the same data splits as TIGEM.

\textbf{SGN.} SGN leverages semantic gene representations to enable zero-shot prediction of unseen genes from histology images. In our implementation, we follow the architecture and training strategy reported in the original work and evaluate SGN under the same SRG 5:1 setting as TIGEM for a direct zero-shot comparison.

\textbf{iStar.} iStar predicts spatial gene expression from histology by leveraging pretrained histology representations and spatial context. In our implementation, we follow the architectural design and training configuration reported in the original work and evaluate the model using the same data splits as TIGEM.

\textbf{mclSTExp.} mclSTExp employs a DenseNet-121 image encoder and a Transformer-based spot encoder with positional encoding to learn a joint embedding space between histology and spatial gene expression through contrastive learning. Gene expression is predicted by similarity-based retrieval and weighted aggregation of the top-$k$ reference expression profiles. In our implementation, we follow the architectural design and training configuration reported in the original work.

\textbf{PVMs.}
We use the official pretrained weights for all PVMs and follow the image preprocessing procedures recommended in their HuggingFace tutorials, including ImageNet mean–std normalization.  
Following the HEST-1K evaluation protocol, all patch embeddings are reduced to 128 dimensions using PCA.  
A Random Forest regressor with 100 trees is then used to predict.

\section{Evaluation Metrics} 
\label{sec:supp_evm}

To assess the quality of spatial expression prediction, we employ three established metrics: the Pearson Correlation Coefficient (PCC), Mean Squared Error (MSE), and Structural Similarity Index Measure (SSIM), together with our proposed Graph Spectral Correlation (GSC). All metrics are computed in the log-transformed expression domain to reduce count-scale variability.

For each gene, PCC evaluates how well the predicted expression captures the directional trend of the ground truth across spatial spots, while MSE reflects the average squared deviation between prediction and measurement. Since the predictive difficulty varies substantially across genes, we follow standard practice~\cite{zhu2025diffusion, chung2024accurate} and summarize PCC at two levels. Specifically, we compute PCC for every gene in a slide, rank all genes by their PCC values, and report the mean PCC over the top 50 and top 200 genes, denoted PCC(50) and PCC(200), respectively.

Formally, given prediction $\hat{y}_{i,j}$ and ground truth $y_{i,j}$ at spot $i$ for gene $j$, PCC and MSE for gene $j$ are defined as
\begin{equation}
    PCC_j = 
\frac{\sum_{i=1}^n ( \hat{y}_{i,j}-\bar{\hat{y}}_{.j} )( y_{i,j}-\bar{y}_{.j} )}
{\sqrt{\sum_{i=1}^n ( \hat{y}_{i,j}-\bar{\hat{y}}_{.j} )^2 }
 \sqrt{\sum_{i=1}^n ( y_{i,j}-\bar{y}_{.j} )^2 }}
\end{equation}
\begin{equation}
    \mathrm{MSE}_j = \frac{1}{n} \sum_{i=1}^n ( y_{i,j} - \hat{y}_{i,j} )^2.
\end{equation}

For a slide with $m$ genes, we then compute:
\begin{equation}
    PCC(50) = \frac{1}{50} \sum_{j \in \text{top-50}} PCC_j,
\end{equation}
\begin{equation}
    PCC(200) = \frac{1}{200} \sum_{j \in \text{top-200}} PCC_j,
\end{equation}
\begin{equation}
    \mathrm{MSE} = \frac{1}{m} \sum_{j=1}^m \mathrm{MSE}_j.
\end{equation}

In addition to these spot-wise metrics, we employ SSIM~\cite{wang2004image} as an established measure of spatial structural similarity. For each gene, the predicted and ground-truth expression values are mapped onto the same two-dimensional spatial lattice, and SSIM evaluates their agreement in local intensity, contrast, and spatial structure. We report SSIM(50) as the mean SSIM over the 50 genes with the highest PCC values:
\begin{equation}
    \mathrm{SSIM}(50)
    =
    \frac{1}{50}
    \sum_{j \in \text{top-50}}
    \mathrm{SSIM}
    \left(
    \hat{\mathbf{Y}}_j,
    \mathbf{Y}_j
    \right),
\end{equation}
where $\hat{\mathbf{Y}}_j$ and $\mathbf{Y}_j$ denote the predicted and ground-truth two-dimensional spatial expression maps for gene $j$, respectively. SSIM is used exclusively for evaluation and is independent of both the TIGEM training objective and the proposed GSC metric.

While PCC and MSE primarily quantify spot-wise numerical agreement, and SSIM provides an established image-based assessment of local spatial structure, they do not explicitly characterize expression patterns with respect to the underlying spatial graph of ST spots. For spatial transcriptomics, such graph-structured organization is particularly relevant for capturing spatial domains and region-specific expression patterns. To complement these established metrics, we introduce GSC, which measures the consistency between predicted and ground-truth expression patterns in the graph spectral domain.

\section{Graph Spectral Correlation}

% TO DO
% 首先调一下布局，按mathematical proof和domain knowledge explanation排版
% 然后在数学证明部分加入和PCC关系的讨论：互补而非替代，PCC就像macro F1一样关注大类别，GSC就像weighted F1一样提升了边界spot（小类别）的权重。
% 加入领域知识，GSC因为K不全为零且不为单位矩阵，假设我们有一个基因，100个spot，这种情况下GSC相对于对于这个基因的空间共表达关系的量化。而PCC则只是在衡量一个单一spot上的对应性。

We propose Graph Spectral Correlation (GSC), a metric that serves as a supplementary for Pearson Correlation Coefficient (PCC). GSC could capture the similarity of the global structure between predicted spatial transcriptomic and ground truth.

Firstly, for one sample, we pick one gene $g$ to measure the quality of prediction. We define the predicted signals as $x_g$ and the ground truth as $y_g$.

PCC is widely used to measure the quality of spatial transcriptomic prediction, defined as:

\begin{equation}
\label{eq:pcc}
    PCC_g = \frac{<x_g, y_g>}{\|x_g\|\,\|y_g\|} =
\frac{x_g^\top I\, y_g}{\|x_g\|\,\|y_g\|}=\frac{x_g^\top I\, y_g}{\sqrt{x_g^\top I\, x_g},\sqrt{y_g^\top I\, y_g}}
\end{equation}

Our GSC is computed by measuring the cosine similarity between the prediction's and the label's spectrum after graph Fourier transform. We first construct the normalized graph Laplacian
\[
L = I - D^{-1/2} A D^{-1/2}
\]
and perform eigendecomposition to obtain the graph Fourier basis. The eigenmodes are ordered according to ascending eigenvalues. From the eigendecomposition of the normalized graph Laplacian, we obtain 
an orthonormal matrix 
\[
U = [u_0, u_1, \ldots, u_{n-1}].
\] 

 We discard the trivial zero-frequency mode and retain the spectral components corresponding to modes \(1\) through \(k\) ($k$ is automatically selected based on the cumulative energy proportion) to compute the final similarity score.

\[
U_k = [u_1, \ldots, u_k].
\] 

Then, for a graph signal \(x\), its truncated graph Fourier spectrum is given by $U_k^\top x$.

Hence, for a gene $g$ and a truncated top $k$ spectrum, our GSC is defined as:

\begin{equation}
    GSC_{g, k}= \frac{<U_k^\top x_g, U_k^\top y_g>}{\|U_k^\top x_g\|\,\|U_k^\top y_g\|} =
    \frac{x_g^\top\, U_k \, U_k^\top \, y_g}{\sqrt{x_g^\top\, U_k \, U_k^\top \, x_g},\sqrt{y_g^\top\, U_k \, U_k^\top \, y_g}}
\end{equation}

Similarly to PCC, the final GSC-measure is the mean value of GSC across genes.

\subsection{Mathematical Proof}
To reveal our GSC's capability to consider the spatial pattern while PCC only evaluates node-wise similarity, we give a simple proof here: \\

We firstly define 
\begin{equation}
    K := U_k U_k^\top = \{K_{ij}\}_{i,j=1}^n
\end{equation}
so the GSC can be rewritten as
\begin{equation}
\begin{aligned}
    GSC_{g,k} 
&= 
\frac{x_g^\top\, K \, y_g}{\sqrt{x_g^\top\, K \, x_g}\sqrt{y_g^\top\, K\, y_g}} \\
&=
\frac{\sum_{i=1}^n \sum_{j=1}^n K_{ij} x_{g,i} y_{g,j}}
{\sqrt{\sum_{i=1}^n \sum_{j=1}^n K_{ij} x_{g,i} x_{g,j}} \;\; \sqrt{\sum_{i=1}^n \sum_{j=1}^n K_{ij} y_{g,i} y_{g,j}}}.
\end{aligned}
\end{equation}

Here, each element of \(K\) is given by
\[
K_{ij} = \sum_{m=1}^{k} u_i^{(m)} u_j^{(m)},
\]
where \(u_i^{(m)}\) denotes the \(i\)-th element of the \(m\)-th eigenvector in \(U_k\). It could also be noted as $U_{k, im}$. \\

% 证明的东西很简单，就是K不是对角矩阵，这样计算GSC的时候，分子就不像是PCC那样算xi--yi, xj--yj了，可能会出现空间位置交叉的情况。但是交叉的空间特性，我觉得很难证明，至少LLM说不能证明K是个什么近大远小核（我觉得就不是，本来这个就是受限的”全局“信息，不知道这里能不能更近一步证明）

% 监督gemini证明的，又找出来不严谨处打了补丁，可以看看是否完备
% 要证的东西不复杂，但是LLM还是蛮费劲的。我发现免费版中gemini比gpt好使

\textbf{Proposition:} $K = U_k U_k^\top$ is not a diagonal matrix, given that:

1. $U_k \in \mathbb{R}^{n \times k}$ consists of $k$ columns selected from an orthonormal matrix $U \in \mathbb{R}^{n \times n}$ ($1 \le k < n-1$) (excluding the DC component $u_0$), where $U$ is the Graph Fourier basis of a connected undirected graph. 

2. $n$ is the number of nodes in the connected undirected graph with adjacency matrix $A$ and degree matrix $D = \text{diag}(d_1, \dots, d_n)$, where $d_i \ge 1$ for all $i \in \{1, \dots, n\}$. \\

\textit{Proof by contradiction}:
\\
Let $r_1^\top, r_2^\top, \dots, r_n^\top$ denote the row vectors of $U_k$. We can represent $U_k$ as
$$U_k = \begin{bmatrix} r_1^\top \\ r_2^\top \\ \vdots \\ r_n^\top \end{bmatrix}$$

Therefore, $K$ can be rewritten by the inner product of its row vectors:
$$K_{ij} = r_i^\top r_j = \langle r_i, r_j \rangle$$

Assume $K$ is a diagonal matrix. This implies that all off-diagonal elements are zero, so we can rewrite as
$$\langle r_i, r_j \rangle = 0 \quad (\forall i \neq j)$$
Under this assumption, the set $\{r_1, r_2, \dots, r_n\}$ must be an orthogonal set of vectors.

Since $\text{rank}(K) = \text{rank}(U_k U_k^\top) = \text{rank}(U_k) = k$, a diagonal matrix with rank $k$ must have exactly $k$ non-zero diagonal elements. Consequently, exactly $k$ rows of $U_k$ are non-zero vectors, and the remaining $n-k$ rows must be zero vectors. Without loss of generality, by reordering the nodes, we can partition $U_k$ as
$$U_k = \begin{bmatrix} V_{k \times k} \\ \mathbf{0}_{(n-k) \times k} \end{bmatrix}$$
where $V$ is a $k \times k$ non-singular submatrix.

%%
%因为是GFT，所以U一定是正交矩阵
Consider the full orthogonal matrix $U = [u_0, U_k, U_{\text{rest}}]$. We partition the constant vector $u_0$ accordingly:$$U = \left[ \begin{array}{c|c|c} u_{0, \text{top}} & V & U_{\text{rest, top}} \\ \hline u_{0, \text{bottom}} & \mathbf{0} & U_{\text{rest, bottom}} \end{array} \right]$$where $u_{0, \text{top}} \in \mathbb{R}^k$. Because the columns of an orthogonal matrix must be mutually orthogonal, $u_0$ must be orthogonal to every column in $U_k$:$$u_0^\top U_k = \mathbf{0}^\top \implies [u_{0, \text{top}}^\top, u_{0, \text{bottom}}^\top] \begin{bmatrix} V \\ \mathbf{0} \end{bmatrix} = \mathbf{0}^\top$$Expanding this:$$u_{0, \text{top}}^\top V = \mathbf{0}^\top$$

% 这里好像也很重要，不然得不到u0上半部分全0，不过u0其实已知了
Since $V$ is a non-singular matrix of rank $k$,$$u_{0, \text{top}} = \mathbf{0}$$
%%
% 这个|ui|公式应该是通用结论，我没有完全查证但是是能推出来的
The result $u_{0, \text{top}} = \mathbf{0}$ implies that the first $k$ components of the constant vector $u_0$ are zero. However, for a connected graph, the unit eigenvector $u_0$ corresponding to the smallest eigenvalue $\lambda_0 = 0$ is unique, and the corresponding eigenvector $u_0$ satisfies
$$L u_0 = 0 \implies D^{-1/2} A D^{-1/2} u_0 = u_0.$$
Writing component wise, for every every pair $(i, j)$ such that $A_{ij} > 0$:

$$\frac{u_{0, i}}{\sqrt{d_i}} = \frac{u_{0, j}}{\sqrt{d_j}} \implies u_{0, i} = c \sqrt{d_i}, \quad \forall i,$$

while $c$ is a constant.  After normalization,  the absolute value of its $i$-th component is given by:$$|u_{0, i}| = \frac{\sqrt{d_i}}{\sqrt{\sum_{j=1}^n d_j}}$$

The degree of each node must satisfy $d_i \ge 1$. Consequently, for every node $i$, we have:$$|u_{0, i}| > 0 \implies u_{0, i} \neq 0$$
This directly contradicts the requirement $u_{0, \text{top}} = \mathbf{0}$ obtained from our assumption.  Hence,  $U_k$ cannot have only $k$ non-zero rows. Similarly, we can also prove that   $U_k$ cannot have only $k^{'} < k$ non-zero rows. The number of non-zero row vectors $m$ must satisfy $m > k$.

In a $k$-dimensional space $\mathbb{R}^k$, the maximum number of mutually orthogonal non-zero vectors is $k$. Since we have $m > k$ non-zero row vectors $r_i$ in $\mathbb{R}^k$, these vectors must be linearly dependent and cannot be mutually orthogonal. Therefore, the assumption is false. There exists at least one pair $i \neq j$ such that $\langle r_i, r_j \rangle \neq 0$, meaning $K$ contains off-diagonal non-zero elements. Thus, $K$ is not a diagonal matrix. $\blacksquare$
\\

Now we have proved that $K$ is not diagonal. Revisiting Eq.~\eqref{eq:pcc}, we can find that PCC's use a $I$ matrix which means for all $i$, PCC's calculation only compares $x_i$ and $y_i$. This preserves the strict matching among spot-pairs, making each spot exactly the same influence to the overall calculated relationship. On the other hand, non-diagonal $K$ brings un-matching pairs $x_i$ and $y_j$ to GSC's calculation. This type of choice enables more spatial-induced weights in computation. Since our spectrum picking from graph Fourier transform keeps low-frequency, which captures relatively large-scale oscillation. The GSC focuses more on large-scale boundaries and borders across regions, which is closely related to the former proved spatial weighted measurement.

Considering the lack of attention on region borders from PCC's universal equal weighting, the proposed GSC could serve as a meaningful complementary metric. It avoids neglecting critical boundaries and local structures on graph as they often get averaged in PCC calculation. Combining these two metrics as a structural quality benchmark could examine both the global numeric relation performance and the significant-regional pattern awareness of models.

%% 总体思路就是： 如果是对角矩阵 --> ri们正交 --> 但是k维的矩阵Uk中不可能n个向量正交 --> 假设冲突
%% 里面有个补丁，{ri, ...}中一定有大于k个非0向量，这样才有效
%% 对角矩阵就保证了一定有xi--yj

%% 后面如果再要写的话，就是说GSC的分子有xi--yj pair，然后分母只局限于自己内部的xi和xj，所以就算是I变K也只是一个scaler，所以GSC的计算是考虑了空间关系的（至于考虑的空间关系是啥样别管）

%%可能要再一步试图证明就是尺度信息的问题了，不过我觉得有点难就是了，毕竟是多个尺度的复合

%% gemini给出的一个思路是：Kij 的大小，数学上严格表现节点 i 和 j 在图拉普拉斯算子下的“局部一致性” 

% 严格的数学表现：距离等价性
% 定义两个节点在谱空间中的平方欧式距离：
% $$d_{\text{spectral}}^2(i, j) = \|r_i - r_j\|^2 = \|r_i\|^2 + \|r_j\|^2 - 2K_{ij}$$由于 $K$ 矩阵是由前 $k$ 个特征向量构造的，根据 Rayleigh Quotient 最小化原理，这个距离 $d_{\text{spectral}}^2(i, j)$ 实际上是在最小化图上的信号突变。
% 数学证明结论：当 $K_{ij}$ 越大，对应于 $d_{\text{spectral}}^2(i, j)$ 越小。这意味着：$K_{ij}$ 的大小，数学上严格表现了节点 $i$ 和 $j$ 在图拉普拉斯算子下的“局部一致性”。 ---

\subsection{Biological Interpretation}

% 空间表达模式（spatial expression pattern）是指单个基因在组织或样本空间中呈现出的系统性分布格局，强调表达在空间上的区域性特征。该概念用于描述基因在组织结构、功能区或微环境中呈现出的整体空间组织方式。空间表达模式主要受组织尺度的空间分区以及发育或病理边界等空间结构性因素影响，具有全局性而非单点属性。因此，传统的PCC只能比较同一空间点的表达值，对局部偏移或噪声敏感，难以衡量基因在整体组织中的空间分布格局。而GSC能够量化基因在整个组织上的空间表达模式一致性，相比 PCC 能够减少局部不匹配带来的信息丢失，更准确地反映基因空间表达的全局特征。

In spatial transcriptomics, expression level is not independent but spatially autocorrelated, forming morphologically distinct clusters of high or low expression. This spatial co-expression phenomenon often leads to different features at the tissue level. Spatial expression pattern refers to the systematic distribution of a single gene across the spatial domain of a tissue or sample. It stresses the regional feature of gene expression in space. This concept is used to describe the overall spatial organization of gene expression shaped by tissue structure, functional compartments, or local microenvironments. Spatial expression patterns are largely shaped by tissue-scale spatial organization such as anatomical compartmentalization and developmental or pathological boundaries, and thus reflect a global property of gene expression rather than a point-wise characteristic.

Considering the overly strict point-wise matching imposed by the definition of PCC across individual sequencing spots, its adaptability to spatial expression patterns is easily weakened by local outliers. With reduced pattern information loss caused by local mismatches, our GSC stresses more in the quantification of spatial gene express pattern in sample level. Therefore, GSC more accurately reflects the global characteristics of spatial gene expression.

\begin{figure}[h!]
  \centering
   \includegraphics[width=0.95\linewidth]{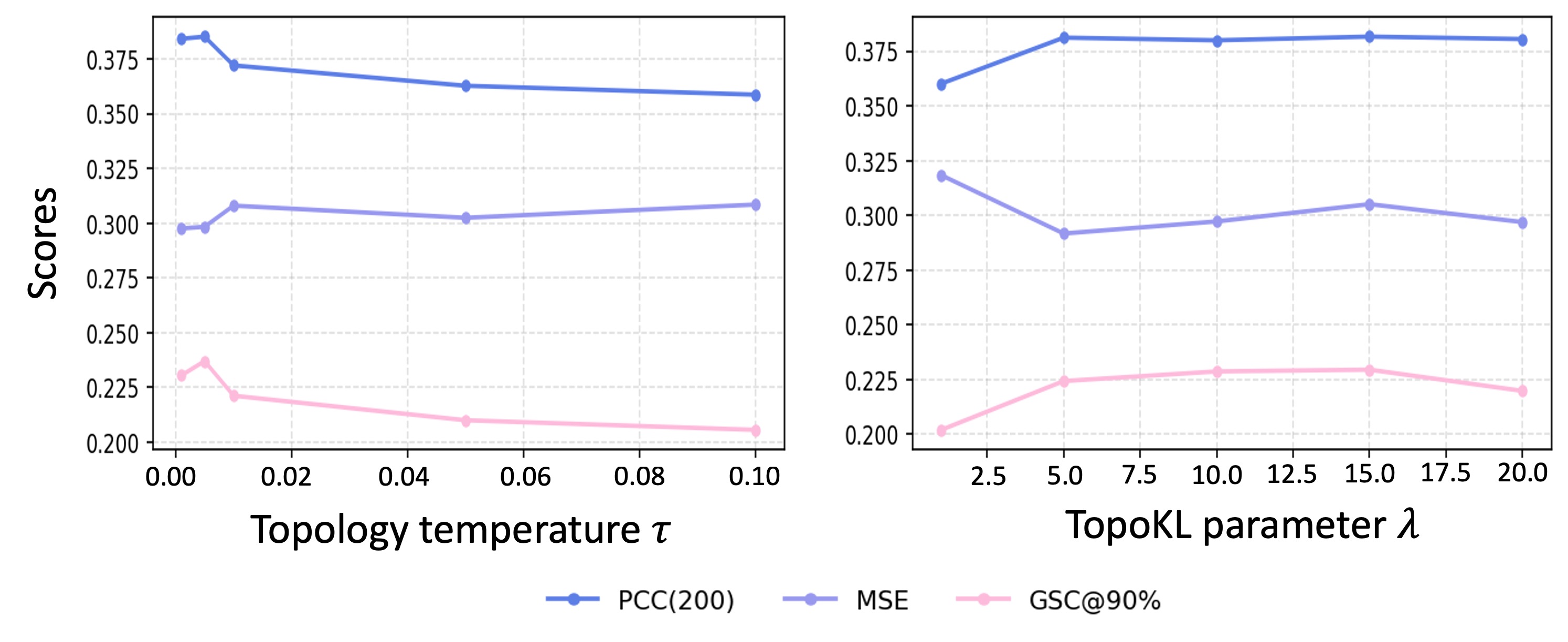}

   \caption{TIGEM performance with different hyperparameter values on the HEART dataset MISC142.}
   \label{fig:six}
\end{figure}

\begin{table*}[h]
\vspace{5pt}
  \caption{Zero-shot comparison between TIGEM and SGN under the SRG 5:1 setting. Best results in each column are highlighted with red backgrounds.}
  \label{tab:zeroshot_sgn}
  \centering
  \small
  \renewcommand{\arraystretch}{0.95}
  \setlength{\tabcolsep}{5.5pt}

  \resizebox{\textwidth}{!}{%
  \begin{tabular}{@{}lccc|ccc|ccc@{}}
    \toprule
    \multirow{2}{*}{Method}
      & \multicolumn{3}{c|}{HEART (MISC142)}
      & \multicolumn{3}{c|}{COAD (MISC36)}
      & \multicolumn{3}{c}{MEND (MEND146)} \\
    \cmidrule(lr){2-4}\cmidrule(lr){5-7}\cmidrule(lr){8-10}
      & MSE $\downarrow$ & PCC(all) $\uparrow$ & GSC@90\% $\uparrow$
      & MSE $\downarrow$ & PCC(all) $\uparrow$ & GSC@90\% $\uparrow$
      & MSE $\downarrow$ & PCC(all) $\uparrow$ & GSC@90\% $\uparrow$ \\
    \midrule

    \textbf{SGN}
      & $0.304_{.003}$ & $0.087_{.007}$ & $0.085_{.003}$ 
      & $1.256_{.002}$ & $0.527_{.004}$ & $0.544_{.005}$
      & $0.435_{.003}$ & $0.266_{.006}$ & $0.284_{.005}$ \\

    \textbf{TIGEM}
      & $0.265_{.005}$ & $0.107_{.004}$ & $0.109_{.005}$
      & $0.973_{.005}$ & $0.639_{.006}$ & $0.687_{.009}$
      & $0.327_{.007}$ & $0.312_{.005}$ & $0.328_{.006}$ \\

    \bottomrule
  \end{tabular}}
  \vspace{-15pt}
\end{table*}

\section{Additional biomarker visualization}

\begin{figure*}[h]
  \centering
  \includegraphics[width=0.95\linewidth]{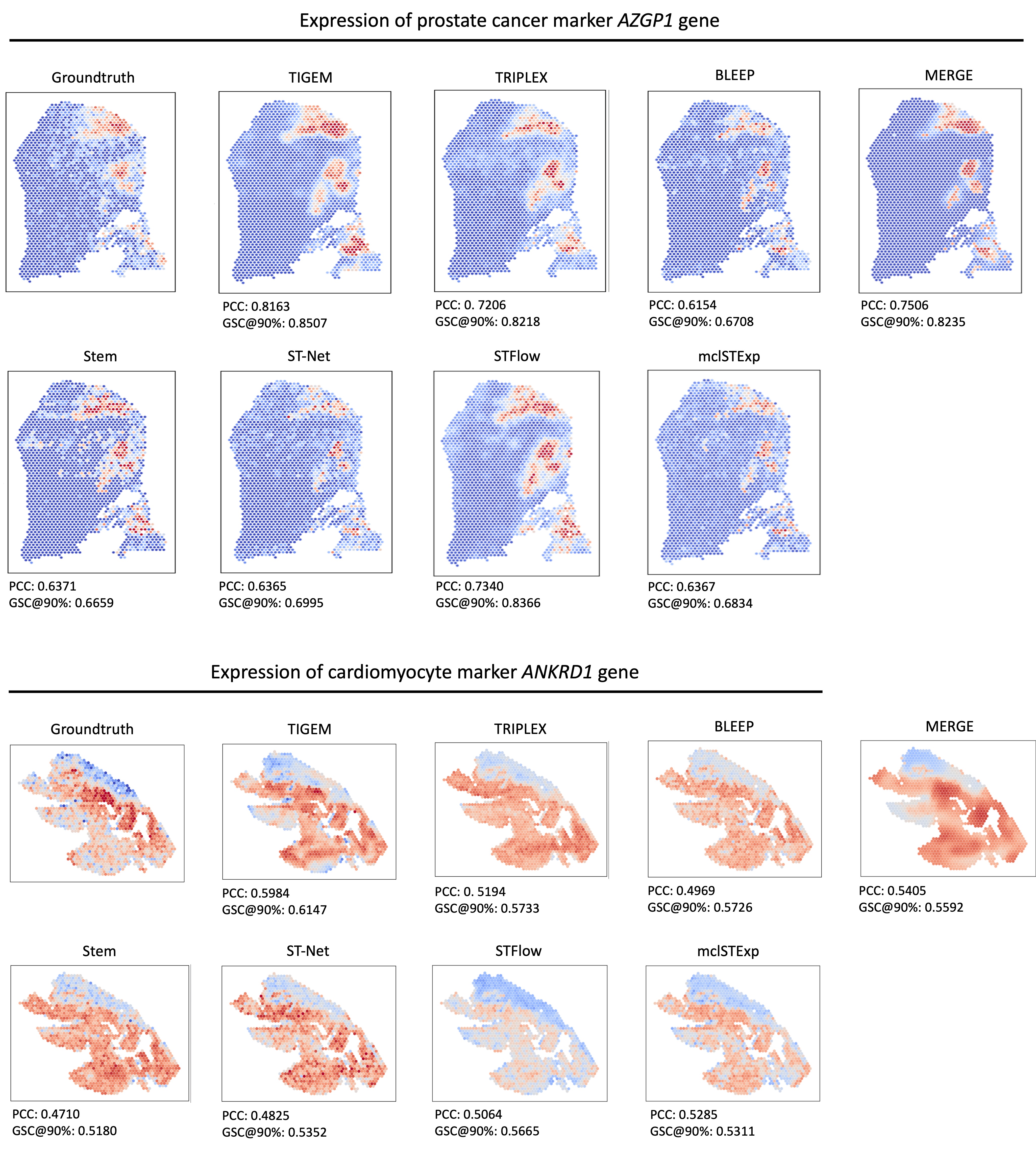}
  \caption{Visualization of biomarker expression prediction across methods. Top shows the prediction of prostate cancer marker AZGP1 on the MEND dataset (MEND146). Bottom shows the prediction of cardiomyocyte marker ANKRD1 on the HEART dataset (MISC142).}
  \label{fig:s1}
\end{figure*}

As we mentioned in main Sec 4.5, we further visualized the cancer marker AZGP1 in the MEND dataset (MEND146) and the cardiomyocyte marker gene ANKRD in the HEART dataset (MISC142). The results are provided in \cref{fig:s1}, which show that TIGEM performs well on different marker genes.

\section{SGN zero-shot prediction benchmarking}

To further evaluate TIGEM against existing zero-shot histology-to-ST methods, we compare it with SGN under the same SRG 5:1 setting across all three datasets. As shown in \cref{tab:zeroshot_sgn}, TIGEM consistently outperforms SGN across all datasets and evaluation metrics. The improvement is particularly pronounced on COAD, where TIGEM increases PCC from $0.527$ to $0.639$ and GSC@90\% from $0.544$ to $0.687$, while reducing MSE from $1.256$ to $0.973$. Consistent improvements are also observed on MEND and HEART. Although the absolute zero-shot performance on HEART remains relatively low for both methods, TIGEM still consistently improves over SGN. These results demonstrate that combining gene-semantic conditioning with topology-aware spatial modeling provides additional benefits for unseen-gene prediction.

\section{Additional HEART dataset in zero-shot evidence experiment}

Here we show the zero-shot evidence experiment on the HEART dataset.

\begin{table}[h]
  \centering
  \scriptsize
  \vspace{-4pt}
  \caption{Justification of the feasibility of strict $G_{\mathrm{test}}\cap G_{\mathrm{train}}=\emptyset$ inference on HEART.}
  \renewcommand{\arraystretch}{1}
  \setlength{\tabcolsep}{4pt}
  \begin{tabular}{@{}lccc@{}} 
    \toprule
    Setting 
    & MSE $\downarrow$ 
    & PCC (All) $\uparrow$ 
    & GSC@90\% $\uparrow$ \\
    \midrule
    SRG 5:1 
    & $0.265_{.005}$ & $0.107_{.004}$ & $0.109_{.005}$ \\
    Random 
    & $0.725_{.003}$ & $0.048_{.007}$ & $0.053_{.004}$ \\
    One-hot 
    & $0.748_{.002}$ & $0.042_{.004}$ & $0.050_{.003}$ \\
    \bottomrule
  \end{tabular}
  \label{tab:supp_fe}
  \vspace{-5pt}
\end{table}
\section{Effect of hyperparameters and model scale}

We next analyzed the impact of key hyperparameters, including the temperature $\tau$, the KL weight $\lambda_{topo}$, and the number of GAT layers $L$ in the TIG module. As shown in \cref{fig:six}, lowering the temperature (sharper topology prior) or increasing $\lambda_{topo}$ consistently improves model performance, supporting the importance of the topology prior for guiding spatial prediction.

We also assessed model scaling by varying the number of GAT layers in the TIG module. As reported in \cref{tab:para}, moderate increases in depth improve accuracy, but performance plateaus once the model reaches 3–5 blocks, indicating limited benefit from further parameter growth.

\begin{table}[h]
  \caption{Effect of TIG block number on spatial gene expression prediction (MISC142).}
  \label{tab:para}
  \centering
  \small
  \renewcommand{\arraystretch}{0.9}
  \setlength{\tabcolsep}{10pt}

  \resizebox{\columnwidth}{!}{%
  \begin{tabular}{@{}cccccc@{}} \toprule
    TIG Block No. & Params. & MSE $\downarrow$ & PCC (50) $\uparrow$ & PCC(200) $\uparrow$ & GSC@90\% $\uparrow$\\ \midrule
      1 & 22M & $0.302_{.002}$ & $0.522_{.009}$ & $0.360_{.015}$ & $0.210_{.005}$ \\
      2 & 26M & $0.300_{.028}$ & $0.532_{.019}$ & $0.371_{.005}$ & $0.223_{.013}$\\
      3 & 30M & $\bestcell{0.291_{.016}}$ & $\bestcell{0.552_{.010}}$ & $\bestcell{0.381_{.024}}$ & $\secondcell{0.230_{.001}}$  \\
      5 & 39M & $\secondcell{0.292_{.021}}$ & $\secondcell{0.549_{.014}}$ & $\secondcell{0.375_{.017}}$ & $\bestcell{0.232_{.013}}$ \\
      10 & 60M & $0.299_{.006}$ & $0.546_{.010}$ & $0.375_{.004}$ & $0.225_{.008}$ \\ \bottomrule
  \end{tabular}}
  \vspace{-0.6em}
\end{table}

\section{Computational Cost Comparison}

\cref{tab:time1}, \cref{tab:time2}, and \cref{tab:time3} provide a detailed comparison of computational cost between TIGEM and all baseline models, measured on a single-slide sample for each fold in each dataset. All timing experiments are run on the Nvidia A40 GPU for fairness. The table reports model parameter counts as well as training and inference times. Training time for TIGEM already includes topology prior calculation and patch embedding generation. Both times are reported in seconds. Training time corresponds to the duration required for each model to reach convergence (or early stopping), while inference time is reported only for Stem, as its DiT-based generative architecture incurs substantial computational overhead during prediction.
\begin{table}[h]
  \caption{Computation cost benchmarking on HEART dataset}
  \label{tab:time1}
  \centering
  \small
  \renewcommand{\arraystretch}{1.05}
  \setlength{\tabcolsep}{12pt}

  \resizebox{\columnwidth}{!}{%
  \begin{tabular}{@{}cccc@{}} \toprule
    Model & \# Params & Training time (sec) & Inference time (sec)\\ \midrule
      ST-Net & 7.5M & $2.12\times10^{4}$&  -  \\
      HistoGene & 73.2M & $6769$ & - \\
      THItoGene & 237M & $6180$ & - \\
      BLEEP & 24.3M & $1.59\times10^{4}$ &  -  \\
      TRIPLEX & 23M & $2.32\times10^{4}$ & -  \\
      Stem & 37.5M & $2.06\times10^5$ & $7.55\times10^4$ \\
      TIGEM & 30M & 5806 & -\\
    
    \bottomrule
  \end{tabular}}
  \vspace{-0.6em}
\end{table}

\begin{table}[h]
  \caption{Computation cost benchmarking on COAD dataset}
  \label{tab:time2}
  \centering
  \small
  \renewcommand{\arraystretch}{1.05}
  \setlength{\tabcolsep}{12pt}

  \resizebox{\columnwidth}{!}{%
  \begin{tabular}{@{}cccc@{}} \toprule
    Model & \# Params & Training time (sec) & Inference time (sec)\\ \midrule
      ST-Net & 7.5M & $2.45\times{10^{4}}$ & - \\
      HistoGene & 73.2M & 3552 & - \\
      THItoGene & 237M & 7095 & - \\
      BLEEP & 24.3M & $3.42\times10^{4}$ & - \\
      TRIPLEX & 23M & $2.96\times10^{4}$ & - \\
      Stem & 37.5M & $2.06\times10^5$ & $1.26\times10^5$ \\
      TIGEM & 30M & 6147 & - \\
    
    \bottomrule
  \end{tabular}}
  \vspace{-0.6em}
\end{table}

\begin{table}[h]
  \caption{Computation cost benchmarking on MEND dataset}
  \label{tab:time3}
  \centering
  \small
  \renewcommand{\arraystretch}{1.05}
  \setlength{\tabcolsep}{12pt}

  \resizebox{\columnwidth}{!}{%
  \begin{tabular}{@{}cccc@{}} \toprule
    Model & \# Params & Training time (sec) & Inference time (sec)\\ \midrule
      ST-Net & 7.5M & $2.21\times10^{4}$ &  -  \\
      HistoGene & 73.2M & 4508 & -  \\
      THItoGene & 237M & 4876 & - \\
      BLEEP & 24.3M & $2.64\times10^{4}$ & -   \\
      TRIPLEX & 23M & $1.82\times10^{4}$ & -  \\
      Stem & 37.5M & $2.07\times10^{5}$ & $6.08\times10^{4}$ \\
      TIGEM & 30M & 6454 & -\\
    
    \bottomrule
  \end{tabular}}
  \vspace{-0.6em}
\end{table}

Benefiting from pre-generated patch embeddings and cached topology priors, TIGEM does not need to train the image feature extractor from scratch thus achieves significantly faster training speed than most of the models across all dataset scales. Notably, Stem requires substantially longer training. Each training takes approximately two to three days, which makes full cross-validation infeasible.

\end{document}